\documentclass{article}
\usepackage{iclr2027_conference,times}

\usepackage{amsmath,amsfonts,bm}

\def\eqref#1{equation~\ref{#1}}

\def\1{\bm{1}}

\def\vd{{\bm{d}}}

\def\vh{{\bm{h}}}

\DeclareMathAlphabet{\mathsfit}{\encodingdefault}{\sfdefault}{m}{sl}
\SetMathAlphabet{\mathsfit}{bold}{\encodingdefault}{\sfdefault}{bx}{n}

\usepackage{hyperref}
\usepackage{url}
\hypersetup{hidelinks}
\usepackage{graphicx}
\usepackage{float}
\usepackage{booktabs}
\usepackage{multirow}
\usepackage[table]{xcolor}
\usepackage{amsmath}
\usepackage{amssymb}
\usepackage{algorithm}
\usepackage{algpseudocode}
\usepackage{placeins}
\usepackage{tabularx}
\usepackage{array}

\title{One Tile, Multiple Instances: Rethinking MIL for Sparse Diagnostic Evidence}

\author{
Runsheng Liu$^{1}$ \quad
Cheng Jin$^{1}$ \quad
Hao Jiang$^{1}$ \quad
Hao Chen$^{1\text{-}6,*}$
\\[2mm]
\parbox{0.94\textwidth}{
\small
\raggedright
$^{1}$Department of Computer Science and Engineering,
The Hong Kong University of Science and Technology,
Hong Kong, China. \\[0.5mm]
$^{2}$Department of Chemical and Biological Engineering,
The Hong Kong University of Science and Technology,
Hong Kong, China. \\[0.5mm]
$^{3}$Division of Life Science,
The Hong Kong University of Science and Technology,
Hong Kong, China. \\[0.5mm]
$^{4}$HKUST Shenzhen-Hong Kong Collaborative Innovation Research Institute,
Futian, Shenzhen, China. \\[0.5mm]
$^{5}$State Key Laboratory of Nervous System Disorders,
The Hong Kong University of Science and Technology,
Hong Kong, China. \\[0.5mm]
$^{6}$SIAT-HKUST Joint Laboratory for Brain Science,
The Hong Kong University of Science and Technology,
Hong Kong, China.
}
\\[1.5mm]
{\small
\texttt{\{rliuar, cheng.jin, hjiangaz\}@connect.ust.hk}
}
\\[0.5mm]
{\small
$^{*}$Corresponding author: \texttt{jhc@ust.hk}
}
}
\newcommand{\res}[2]{#1\,{\tiny(#2)}}
\newcommand{\ours}[1]{\cellcolor{gray!7}#1}
\newcommand{\oursres}[2]{\cellcolor{gray!7}\res{#1}{#2}}
\newcommand{\greyrow}{\rowcolor{gray!10}}

\newcommand{\pdelta}[1]{\textcolor{green!50!black}{+#1}}
\newcommand{\ndelta}[1]{\textcolor{red!70!black}{-#1}}
\iclrfinalcopy 
\begin{document}

\maketitle
\lhead{arXiv preprint}
\begin{abstract}
In weakly supervised Whole Slide Image (WSI) classification, feature extractors typically compress each image tile into a single global embedding. Consequently, slide-level aggregators are restricted to this coarse tile scale, concealing fine-grained sub-tile evidence from the attention mechanism. We introduce DI-MIL, a framework that decouples encoding context from instance granularity through decomposed instances. By clustering dense spatial tokens from a frozen foundation model within each tile, DI-MIL converts a single tile into multiple independently weighted instance embeddings. As a training-free post-encoding module, DI-MIL integrates seamlessly into existing pipelines without requiring re-encoding or downstream architectural modifications. We evaluate DI-MIL on cytopathology, a challenging testbed where sparse diagnostic signals are easily diluted within standard tiles. Across four datasets, three frozen foundation models, and two attention-based aggregators, DI-MIL demonstrates consistent efficacy, improving 67 of 72 metric-level comparisons, with the largest mean gains reaching 3.64 points under cytopathology-specific backbones. In a broader comparison against seven representative MIL baselines, DI-MIL paired with ACMIL achieves highest mean performance in 33 of 36 backbone-dataset-metric comparisons. Ablations show that direct smaller tiling inflates the extracted tile count by up to 43.3$\times$ with non-monotonic performance, whereas DI-MIL incurs zero additional image-extraction overhead while achieving the strongest overall results. These results establish instance construction as an orthogonal design dimension in MIL, supporting DI-MIL as a cost-efficient solution under sparse diagnostic evidence. Code will be released upon acceptance.
\end{abstract}

\section{Introduction}
\label{sec:introduction}

Deep learning-based pathology diagnosis on whole-slide images (WSIs) presents a classic weak-supervision problem, where sparse visual evidence determines the slide-level label. Multiple instance learning (MIL) addresses this by treating a slide as a bag of instances aggregated, typically via attention mechanisms under slide-level supervision \citep{ilse2018attention,li2021dsmil,shao2021transmil}. Typically, standard MIL workflows adopt a two-stage paradigm: image tiles are extracted from WSIs and encoded into fixed feature embeddings by a pre-trained backbone, which are subsequently aggregated into a slide-level representation. This standard pipeline compresses each tile into a single global embedding, rigidly coupling aggregation granularity to the feature extractor's receptive field and forcing the model to treat the tile as indivisible.

However, this coupling poses a limitation when diagnostic evidence is sparse within a tile. Cytopathology provides a concrete example: scattered malignant cells are often interspersed among abundant normal cells and non-diagnostic background \citep{dov2021weakly,butke2021end,zhao2024less}. In such cases, a tile can be diagnostically decisive even if only a minute fraction of its region is relevant. Because each tile is represented as a single embedding, attention mechanisms can only assign a single weight to the tile as a whole. As a result, slide attention cannot weight critical sub-tile morphology independently of surrounding background, conflating tile identification with sub-tile evidence weighting.

To address this, we argue that the spatial encoding context of the feature extractor should be decoupled from the instance granularity used in slide-level aggregation. A straightforward workaround is to extract smaller source tiles; however, this presents an inherent trade-off: while smaller tiles increase aggregation granularity, they severely shrink the spatial context provided to the backbone. Vision foundation models offer a way out: their dense spatial tokens preserve fine-grained localized semantics while maintaining a broad contextual field of view \citep{simeoni2025dinov3,li2025pathvq}. This capability challenges the traditional MIL assumption, raising our core question: \emph{must features encoded together within a tile also be aggregated as a single instance?} We posit that the source tile should remain the unit of contextual encoding, while its sub-tile spatial tokens serve as individual instances for slide-level aggregation.

To this end, we introduce \textbf{DI-MIL}, a feature-level framework that achieves this separation via decomposed instances. For each tile, DI-MIL clusters \(\ell_2\)-normalized spatial tokens by feature similarity, averages the original token representations within each cluster, and applies layer normalization to the pooled cluster embeddings. Each cluster embedding enters the slide bag as an independent instance, allowing sub-tile components to receive distinct attention weights. By expanding a slide bag of $N$ source tiles into $kN$ fine-grained instances, DI-MIL enriches instance granularity without requiring additional image crops or extra encoder forward passes.

Using cytopathology as a testbed for sparse evidence, we systematically evaluate DI-MIL as a plug-and-play enhancement across a full matrix of 4 datasets, 3 vision foundation backbones, and 2 MIL aggregators. Across 72 total evaluation settings combining multiple datasets, backbones, aggregators, and metrics, DI-MIL improves performance in 67 configurations, where maximum mean gains reaching 3.64 points under cytopathology-specific backbones. In a broader comparison against seven representative MIL baselines, DI-MIL combined with ACMIL achieves the highest mean result in 33 of 36 backbone-dataset-metric comparisons. Ablation studies establish $k=4$ as a default setting and demonstrate that extracting smaller source tiles substantially increases computational and storage overhead without matching the performance achieved by DI-MIL.

In summary, our main contributions are threefold:
\begin{itemize}
\item We delineate the distinction between encoding context and instance granularity in MIL, uncovering the structural limitation of single-embedding tiles when diagnostic evidence is sparse within a tile.
\item We propose DI-MIL, a framework that decomposes dense spatial tokens into independently weighted instances, refining aggregation granularity without requiring extra encoder forward passes or cell-level annotations.
\item We extensively evaluate DI-MIL on cytopathology benchmarks, demonstrating that decoupling encoding context from instance granularity outperforms both standard MIL baselines and naive extraction of smaller source tiles.
\end{itemize}
\section{Related Work}
\label{sec:related}

\paragraph{Aggregation in multi-instance learning.}
Multiple instance learning (MIL) algorithms for whole-slide image (WSI) classification have primarily focused on downstream aggregation mechanisms, including attention pooling \citep{ilse2018attention}, instance-relation modeling \citep{li2021dsmil,shao2021transmil}, pseudo-bag distillation \citep{zhang2022dtfd}, and long-sequence modeling \citep{yang2024mambamil}. Subsequent advances such as ACMIL and ASMIL refine attention calibration during pooling \citep{zhang2024acmil,ye2026asmil}, while HMIL introduces hierarchical alignment across scales \citep{jin2024hmil}. While existing aggregators innovate in feature grouping and weighting, they treat each tile as a monolithic vector. DI-MIL decomposes dense spatial tokens into fine-grained instances prior to slide-level aggregation, making it fully complementary to downstream pooling architectures.

\paragraph{Dense representations in WSI learning.}
Providing such fine-grained inputs relies on dense feature representations. Modern vision foundation models extract dense token features alongside global image representations, providing fine-grained spatial information \citep{simeoni2025dinov3}. In computational pathology, methods such as PathVQ retain dense spatial tokens via vector quantization \citep{li2025pathvq}. However, PathVQ compresses intra-tile spatial tokens into a single tile representation prior to slide-level pooling. In contrast, DI-MIL preserves multiple cluster embeddings per tile as separate inputs throughout the MIL aggregation stage to prevent localized pathology from being diluted prior to slide-level weighting.

\paragraph{Token clustering and prototype construction.}
To structure and condense such dense spatial tokens, token clustering is widely employed. Organizing spatial representations through clustering has been explored across various stages of model pipelines. For instance, TCFormer progressively merges tokens inside the visual encoder to adapt spatial receptive fields \citep{zeng2022not}. Conversely, PTCMIL applies prompt-guided clustering over global tile representations across the whole slide to construct slide-level prototypes \citep{zhao2026ptcmil}. Distinct from these approaches, DI-MIL performs intra-tile spherical clustering independently on dense tokens after frozen feature extraction. This choice dynamically reduces token redundancy while strictly preserving local spatial context, avoiding the computational complexity of slide-wide clustering.

\paragraph{Cytopathology and sparse diagnostic evidence.}
The need for sub-tile instance decomposition is most critical in cytopathology, where evidence sparsity is extreme. Weakly supervised learning in cytopathology addresses severe evidence sparsity across various clinical applications, including thyroid malignancy prediction \citep{dov2021weakly}, urothelial carcinoma classification \citep{butke2021end}, and malignant effusion classification \citep{kim2025effusion}. Domain-specific architectures such as LESS address irregular cell distributions via multi-scale modeling \citep{zhao2024less}, while frameworks such as Smart-CCS and UniCAS leverage specialized foundation model pretraining \citep{jiang2025smartccs,jiang2026unicas}. These works establish cytopathology as a benchmark for learning under sparse diagnostic evidence. We utilize this setting to evaluate how intra-tile dense tokens should be structured into MIL instances when informative content represents only a fraction of the tile area.

\section{Decomposed-Instance Multiple Instance Learning}
\label{sec:method}

Conventional WSI multiple-instance learning (MIL) uses the encoder tile as the basic unit of slide-level aggregation. This couples two distinct design choices: the spatial context seen by the image encoder and the granularity at which the MIL model can weight diagnostic evidence. Using smaller tiles can produce a finer-grained bag, but simultaneously changes the encoder input and increases the number of image crops and encoder evaluations.

\begin{figure}[!t]
    \centering
    \includegraphics[width=0.92\textwidth]{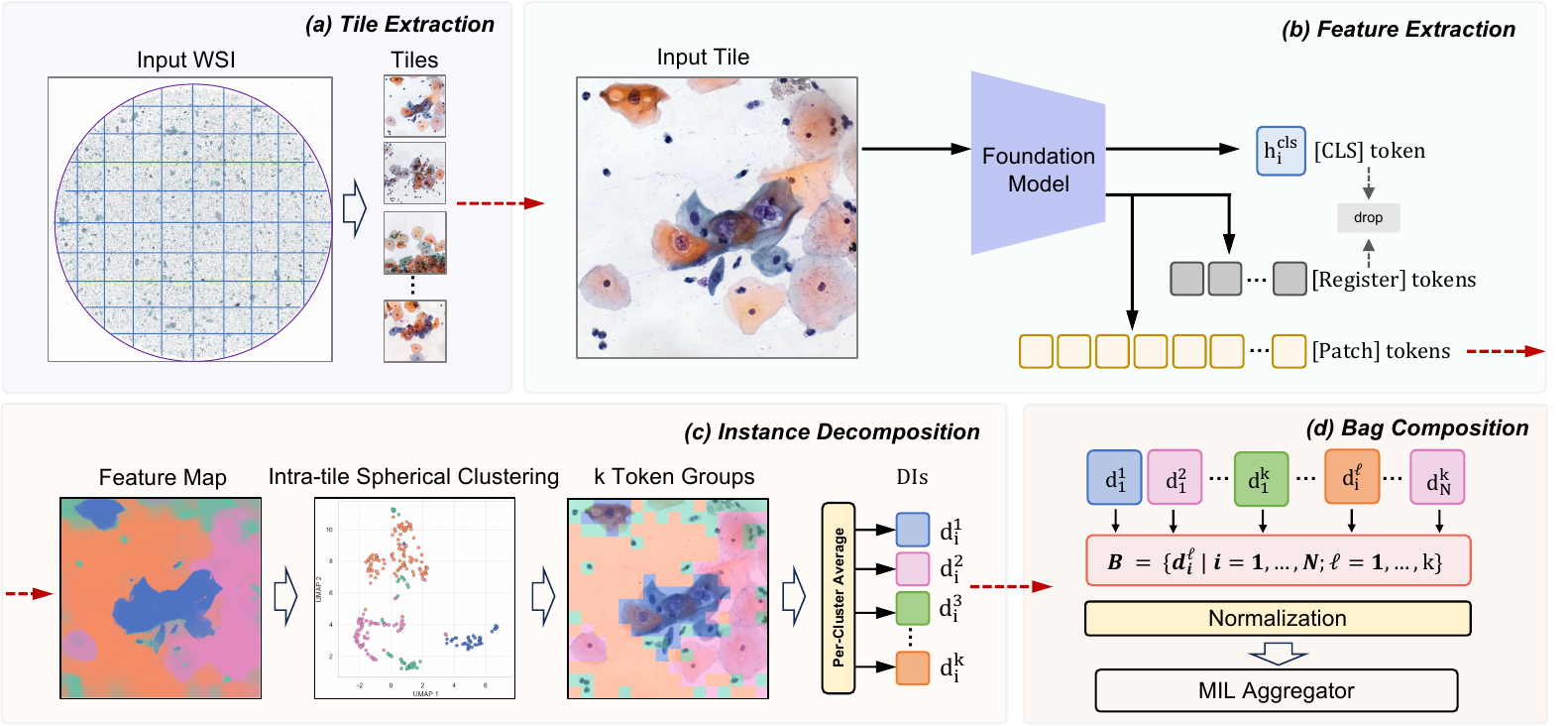}
    \caption{\textbf{Overview of DI-MIL.}
    \textbf{(a)} A whole-slide cytopathology image is partitioned into source tiles.
    \textbf{(b)} A frozen transformer-based foundation model extracts a global [CLS] representation and dense patch-token features from each tile.
    \textbf{(c)} DI-MIL performs intra-tile spherical clustering on dense token representations, aggregates tokens within each cluster, and constructs multiple decomposed instances per tile.
    \textbf{(d)} The resulting decomposed instances from all tiles are assembled into the slide-level MIL bag and processed by a downstream MIL aggregator.
    }
    \label{fig:method}
\end{figure}

We introduce \textbf{Decomposed-Instance MIL (DI-MIL)} to decouple these two factors. DI-MIL preserves the original tile extraction and encoder field of view, while decomposing the patch-token representations already produced for each tile into multiple MIL instances. The slide-level model can therefore operate at a finer instance granularity without re-encoding the WSI at a smaller tile size. An overview is shown in Figure~\ref{fig:method}.

\vspace{-4pt}
\vspace{-4pt}

\subsection{Preliminaries}
\label{sec:formulation}

Let a whole-slide image $X$ be partitioned into $N$ non-overlapping image tiles $\{p_i\}_{i=1}^{N}$. A frozen transformer encoder $f$ processes each tile independently. We denote the final-layer class-token representation by $\vh_i^{\mathrm{cls}}\in\mathbb{R}^{D}$ and the corresponding $M$ patch-token representations by $\{\vh_{ij}\}_{j=1}^{M}$, where $\vh_{ij}\in\mathbb{R}^{D}$. Thus,
\begin{equation}
    f(p_i)
    =
    \left(
        \vh_i^{\mathrm{cls}},
        \{\vh_{ij}\}_{j=1}^{M}
    \right).
    \label{eq:encoder}
\end{equation}

Conventional WSI MIL retains only the class-token representation of each tile, forming the slide-level bag
\begin{equation}
    \mathcal{B}_{\mathrm{tile}}
    =
    \{\vh_i^{\mathrm{cls}}\}_{i=1}^{N},
    \qquad
    \hat{y}
    =
    g_{\theta}(\mathcal{B}_{\mathrm{tile}}),
    \label{eq:tile-bag}
\end{equation}
where $g_{\theta}$ is an attention-based MIL aggregator.

Equating the MIL instance to the tile-level embedding $\vh_i^{\mathrm{cls}}$ restricts instance granularity to the physical tile. Consequently, visual patterns that are distinct within the same tile cannot receive separate MIL attention weights, although they remain represented by different patch-token features within the encoder output.

We refer to the amount of encoded visual content summarized by one independently weighted MIL instance as \emph{instance granularity}. Conventional MIL fixes this granularity at the tile level. DI-MIL instead constructs multiple MIL instances from the patch-token representations of each tile while keeping the encoder input unchanged.

\subsection{Decomposed-instance construction}
\label{sec:di}

In cytopathology, an extracted tile $p_i$ exhibits severe intra-tile heterogeneity, often blending sparse diagnostically abnormal cells with abundant benign background. Compressing $p_i$ into a single global embedding $\vh_i^{\mathrm{cls}}$ forces the aggregator to assign a single attention weight to this mixed representation, diluting localized diagnostic evidence. By contrast, for each tile $p_i$, DI-MIL replaces its single tile-level representation $\vh_i^{\mathrm{cls}}$ with $k$ decomposed instances constructed from the corresponding patch-token representations $\{\vh_{ij}\}_{j=1}^{M}$. As illustrated in Fig.~\ref{fig:method}, while the frozen encoder outputs both [CLS] token and patch tokens, DI-MIL bypasses the global class-token and restructures only the patch-token representations into slide-level MIL instances without altering the upstream forward pass.

\paragraph{Intra-tile token clustering.}
For each tile $p_i$, we group its $M$ patch-token representations by cosine similarity in feature space. We first $\ell_2$-normalize each representation as $\bar{\vh}_{ij} = \vh_{ij}/\lVert\vh_{ij}\rVert_2$ and apply spherical $k$-means independently per tile. Denoting $\boldsymbol{\mu}_{i\ell}$ as the unit-norm centroid of the $\ell$-th cluster, each patch representation is assigned to its nearest centroid:
\begin{equation}
    a_{ij} = \arg\max_{\ell\in\{1,\ldots,k\}} \bar{\vh}_{ij}^{\top}\boldsymbol{\mu}_{i\ell},
    \qquad
    \mathcal{C}_{i\ell} = \{j \mid a_{ij}=\ell\},
    \qquad
    \lVert\boldsymbol{\mu}_{i\ell}\rVert_2=1.
    \label{eq:token-clustering}
\end{equation}
This yields $k$ disjoint token groups $\{\mathcal{C}_{i1},\ldots,\mathcal{C}_{ik}\}$ for each tile. Because clustering is performed independently in feature space, it requires neither spatial contiguity among patches within a cluster nor cross-tile cluster alignment.

\paragraph{Decomposed-instance pooling.}
Each cluster is converted into a decomposed instance by averaging the original patch representations assigned to it, followed by layer normalization:
\begin{equation}
    \vd_i^{\ell} = \operatorname{LN}\!\left( \frac{1}{|\mathcal{C}_{i\ell}|} \sum_{j\in\mathcal{C}_{i\ell}} \vh_{ij} \right), \qquad \ell=1,\ldots,k.
    \label{eq:cluster-pool}
\end{equation}
Note that the normalized representations $\bar{\vh}_{ij}$ are used strictly for cluster assignment, whereas the final instances $\vd_i^{\ell}$ are pooled from the original encoder representations $\vh_{ij}$. Consequently, each tile contributes $k$ decomposed instances $\{\vd_i^{\ell}\}_{\ell=1}^{k}$ instead of a single global vector, yielding the slide-level DI bag across all $N$ tiles:
\begin{equation}
    \mathcal{B}_{\mathrm{DI}} = \left\{ \vd_i^{\ell} \mid i=1,\ldots,N,\; \ell=1,\ldots,k \right\}, \qquad |\mathcal{B}_{\mathrm{DI}}|=kN.
    \label{eq:di-bag}
\end{equation}
In effect, DI-MIL enriches the instance granularity exposed to slide-level aggregation while leaving both the source image tiles and the upstream visual encoding computation entirely untouched.

\subsection{MIL Integration and Computational Overhead}
\label{sec:integration}

By aforementioned construction, each decomposed instance $\vd_i^{\ell}$ shares the same feature dimensionality as the original tile-level representation $\vh_i^{\mathrm{cls}}$. This allows the expanded bag $\mathcal{B}_{\mathrm{DI}}$ to be seamlessly processed by standard attention-based MIL aggregators without any architectural modifications:
\begin{equation}
    \hat{y} = g_{\theta}(\mathcal{B}_{\mathrm{DI}}).
    \label{eq:di-pred}
\end{equation}
For permutation-invariant aggregators, the local cluster index $\ell$ requires no cross-tile alignment; each $\vd_i^{\ell}$ is simply treated as an independent instance within the slide-level bag.

In summary, DI-MIL alters instance granularity purely at the feature level, expanding the slide-level bag size from $N$ to $kN$ without modifying the image-encoding stage or requiring additional forward passes. Given pre-extracted tokens, decomposed instances are constructed once and cached prior to training, incurring minimal computational overhead compared to visual re-encoding.
\section{Experiments}
\label{sec:experiments}

\subsection{Experimental Setup}
\label{sec:setup}

We evaluate four binary cytopathology WSI datasets: \textit{Cervical} (\(978\) slides), \textit{Serous Fluid} (\(2{,}717\) slides), \textit{Respiratory} (\(1{,}041\) slides), and the public \textit{Oral2026} dataset (\(363\) slides) \citep{jain2026oral} released on Zendono.\footnote{\url{https://doi.org/10.5281/zenodo.20643817}}. For the three institutional cohorts, non-overlapping \(1200\times1200\)-pixel tiles are extracted after Otsu-based background filtering. \textit{Oral2026} uses \(2048\times2048\)-pixel level-0 tiles following the established protocol. The institutional cohorts use fixed 10-fold cross-validation, whereas \textit{Oral2026} uses 5-fold cross-validation following previous work \citep{mukherjee2025raamilnovelframeworkclassification}.
We report metrics across folds including AUC, accuracy (ACC), and macro-F1. 

\vspace{-4pt}
\begin{table}[!ht]
\centering
\caption{
Comparison between Tile instances and our Decomposed instances (DI) across 4 cytopathology datasets, 3 foundation-model backbones, and 2 MIL aggregators.
All methods use identical downstream MIL architectures and training protocols.
Results are reported as mean with standard deviation in parentheses.
$\Delta$ denotes the absolute performance change from Tile to DI.
}
\label{tab:main_results}

\scriptsize
\setlength{\tabcolsep}{6.0pt}
\renewcommand{\arraystretch}{1.08}

\begin{tabular}{ll|ccc|ccc}
\toprule
\multirow{2}{*}{Dataset}
& \multirow{2}{*}{Instance}
& \multicolumn{3}{c|}{ABMIL}
& \multicolumn{3}{c}{ACMIL} \\
\cmidrule(lr){3-5}
\cmidrule(lr){6-8}
&
& AUC & ACC & F1
& AUC & ACC & F1 \\
\midrule


\multicolumn{8}{c}{\textbf{DINOv3}} \\
\midrule

\multirow{3}{*}{Cervical}
& Tile
& \res{91.40}{2.11}
& \res{84.56}{3.45}
& \res{84.40}{3.44}
& \res{93.83}{1.96}
& \res{87.22}{3.33}
& \res{87.01}{3.43} \\

& \ours{DI (Ours)}
& \oursres{91.52}{2.65}
& \oursres{85.69}{3.46}
& \oursres{85.48}{3.59}
& \oursres{93.86}{1.97}
& \oursres{88.24}{3.12}
& \oursres{88.06}{3.23} \\

& $\Delta$
& \pdelta{0.12}
& \pdelta{1.13}
& \pdelta{1.08}
& \pdelta{0.03}
& \pdelta{1.02}
& \pdelta{1.05} \\
\cmidrule(lr){1-8}

\multirow{3}{*}{Serous Fluid}
& Tile
& \res{85.37}{2.57}
& \res{78.76}{2.23}
& \res{78.70}{2.27}
& \res{86.11}{3.49}
& \res{79.13}{3.21}
& \res{79.11}{3.21} \\

& \ours{DI (Ours)}
& \oursres{85.80}{3.58}
& \oursres{79.53}{3.45}
& \oursres{79.49}{3.49}
& \oursres{86.79}{3.50}
& \oursres{80.09}{3.56}
& \oursres{80.06}{3.56} \\

& $\Delta$
& \pdelta{0.43}
& \pdelta{0.77}
& \pdelta{0.79}
& \pdelta{0.68}
& \pdelta{0.96}
& \pdelta{0.95} \\
\cmidrule(lr){1-8}

\multirow{3}{*}{Respiratory}
& Tile
& \res{79.38}{4.18}
& \res{74.73}{4.14}
& \res{70.64}{5.31}
& \res{81.05}{4.66}
& \res{75.98}{3.15}
& \res{72.11}{3.87} \\

& \ours{DI (Ours)}
& \oursres{80.77}{5.35}
& \oursres{76.37}{4.50}
& \oursres{71.58}{7.91}
& \oursres{82.01}{4.92}
& \oursres{76.75}{4.19}
& \oursres{72.67}{5.70} \\

& $\Delta$
& \pdelta{1.39}
& \pdelta{1.64}
& \pdelta{0.94}
& \pdelta{0.96}
& \pdelta{0.77}
& \pdelta{0.56} \\
\cmidrule(lr){1-8}

\multirow{3}{*}{Oral2026}
& Tile
& \res{88.06}{5.07}
& \res{79.84}{2.85}
& \res{70.93}{5.04}
& \res{88.37}{5.15}
& \res{80.77}{2.89}
& \res{71.45}{5.90} \\

& \ours{DI (Ours)}
& \oursres{85.72}{6.30}
& \oursres{82.61}{3.01}
& \oursres{66.26}{12.28}
& \oursres{89.36}{4.53}
& \oursres{79.91}{4.09}
& \oursres{72.47}{5.20} \\

& $\Delta$
& \ndelta{2.34}
& \pdelta{2.77}
& \ndelta{4.67}
& \pdelta{0.99}
& \ndelta{0.86}
& \pdelta{1.02} \\
\cmidrule(lr){1-8}


\multicolumn{8}{c}{\textbf{Smart-CCS}} \\
\midrule

\multirow{3}{*}{Cervical}
& Tile
& \res{94.38}{1.58}
& \res{87.42}{2.41}
& \res{87.28}{2.53}
& \res{94.31}{1.90}
& \res{86.82}{3.36}
& \res{86.61}{3.48} \\

& \ours{DI (Ours)}
& \oursres{94.71}{1.95}
& \oursres{90.08}{1.93}
& \oursres{89.99}{1.95}
& \oursres{95.89}{1.21}
& \oursres{89.98}{3.36}
& \oursres{89.87}{3.35} \\

& $\Delta$
& \pdelta{0.33}
& \pdelta{2.66}
& \pdelta{2.71}
& \pdelta{1.58}
& \pdelta{3.16}
& \pdelta{3.26} \\
\cmidrule(lr){1-8}

\multirow{3}{*}{Serous Fluid}
& Tile
& \res{84.22}{2.22}
& \res{76.52}{2.74}
& \res{76.49}{2.74}
& \res{85.04}{2.82}
& \res{77.51}{3.20}
& \res{77.43}{3.25} \\

& \ours{DI (Ours)}
& \oursres{87.63}{2.85}
& \oursres{80.05}{2.54}
& \oursres{79.98}{2.61}
& \oursres{88.33}{2.09}
& \oursres{80.64}{2.62}
& \oursres{80.60}{2.63} \\

& $\Delta$
& \pdelta{3.41}
& \pdelta{3.53}
& \pdelta{3.49}
& \pdelta{3.29}
& \pdelta{3.13}
& \pdelta{3.17} \\
\cmidrule(lr){1-8}

\multirow{3}{*}{Respiratory}
& Tile
& \res{79.36}{4.13}
& \res{74.16}{3.11}
& \res{70.04}{4.38}
& \res{81.21}{3.45}
& \res{76.17}{3.19}
& \res{71.83}{4.47} \\

& \ours{DI (Ours)}
& \oursres{87.50}{2.30}
& \oursres{80.11}{2.14}
& \oursres{77.25}{2.79}
& \oursres{88.85}{2.07}
& \oursres{82.32}{3.10}
& \oursres{80.08}{3.48} \\

& $\Delta$
& \pdelta{8.14}
& \pdelta{5.95}
& \pdelta{7.21}
& \pdelta{7.64}
& \pdelta{6.15}
& \pdelta{8.25} \\
\cmidrule(lr){1-8}

\multirow{3}{*}{Oral2026}
& Tile
& \res{83.91}{5.38}
& \res{83.18}{2.74}
& \res{67.18}{13.01}
& \res{88.20}{5.44}
& \res{81.73}{6.14}
& \res{72.26}{4.25} \\

& \ours{DI (Ours)}
& \oursres{84.17}{6.64}
& \oursres{81.88}{3.13}
& \oursres{71.26}{6.45}
& \oursres{89.13}{4.86}
& \oursres{84.52}{5.25}
& \oursres{75.69}{8.27} \\

& $\Delta$
& \pdelta{0.26}
& \ndelta{1.30}
& \pdelta{4.08}
& \pdelta{0.93}
& \pdelta{2.79}
& \pdelta{3.43} \\
\cmidrule(lr){1-8}


\multicolumn{8}{c}{\textbf{UniCAS}} \\
\midrule

\multirow{3}{*}{Cervical}
& Tile
& \res{93.68}{2.43}
& \res{88.66}{2.36}
& \res{88.54}{2.39}
& \res{95.42}{0.74}
& \res{89.06}{1.73}
& \res{88.84}{1.85} \\

& \ours{DI (Ours)}
& \oursres{94.35}{1.19}
& \oursres{89.27}{2.35}
& \oursres{89.15}{2.40}
& \oursres{95.79}{1.62}
& \oursres{89.57}{2.14}
& \oursres{89.39}{2.23} \\

& $\Delta$
& \pdelta{0.67}
& \pdelta{0.61}
& \pdelta{0.61}
& \pdelta{0.37}
& \pdelta{0.51}
& \pdelta{0.55} \\
\cmidrule(lr){1-8}

\multirow{3}{*}{Serous Fluid}
& Tile
& \res{84.04}{2.21}
& \res{76.85}{2.04}
& \res{76.81}{2.01}
& \res{84.45}{2.40}
& \res{76.52}{2.27}
& \res{76.48}{2.26} \\

& \ours{DI (Ours)}
& \oursres{85.13}{3.09}
& \oursres{79.10}{2.69}
& \oursres{79.07}{2.70}
& \oursres{86.06}{2.41}
& \oursres{78.43}{2.04}
& \oursres{78.39}{2.01} \\

& $\Delta$
& \pdelta{1.09}
& \pdelta{2.25}
& \pdelta{2.26}
& \pdelta{1.61}
& \pdelta{1.91}
& \pdelta{1.91} \\
\cmidrule(lr){1-8}

\multirow{3}{*}{Respiratory}
& Tile
& \res{75.92}{4.99}
& \res{73.78}{3.39}
& \res{69.89}{4.28}
& \res{78.70}{2.84}
& \res{74.54}{2.70}
& \res{69.63}{3.91} \\

& \ours{DI (Ours)}
& \oursres{80.77}{5.48}
& \oursres{74.54}{5.38}
& \oursres{70.31}{6.85}
& \oursres{80.73}{4.05}
& \oursres{75.70}{4.04}
& \oursres{71.41}{4.46} \\

& $\Delta$
& \pdelta{4.85}
& \pdelta{0.76}
& \pdelta{0.42}
& \pdelta{2.03}
& \pdelta{1.16}
& \pdelta{1.78} \\
\cmidrule(lr){1-8}

\multirow{3}{*}{Oral2026}
& Tile
& \res{85.92}{4.96}
& \res{85.30}{7.65}
& \res{71.74}{8.80}
& \res{87.62}{3.40}
& \res{84.82}{1.27}
& \res{71.63}{6.63} \\

& \ours{DI (Ours)}
& \oursres{88.65}{2.75}
& \oursres{86.18}{2.40}
& \oursres{77.14}{5.57}
& \oursres{88.76}{3.34}
& \oursres{84.60}{2.40}
& \oursres{74.38}{5.66} \\

& $\Delta$
& \pdelta{2.73}
& \pdelta{0.88}
& \pdelta{5.40}
& \pdelta{1.14}
& \ndelta{0.22}
& \pdelta{2.75} \\

\bottomrule
\end{tabular}

\end{table}
\vspace{-8pt}

\paragraph{Backbones, MIL aggregators, and training.}
Feature extraction uses one general-domain foundation model, DINOv3 \citep{simeoni2025dinov3}, alongside two cytopathology-specific foundation models, Smart-CCS \citep{jiang2025smartccs} and UniCAS \citep{jiang2026unicas}. Evaluations apply DI-MIL directly to two set-based attention aggregators: ABMIL \citep{ilse2018attention} and ACMIL \citep{zhang2024acmil}. Sequence-based and spatial-structure MIL architectures serve as external reference baselines. All experiments use frozen 1024-dimensional embeddings, a learning rate of \(2\times10^{-4}\), a batch size of one WSI, and early stopping within 100 epochs. DI-MIL uses \(k=4\) by default. Within each paired comparison, the extracted features, data splits, and training protocol are held fixed. Full settings are provided in Appendix~\ref{app:setup_details}.

\subsection{Main Result: Performance across Backbones and Aggregators}
\label{sec:main_result}

Table~\ref{tab:main_results} summarizes the performance of DI-MIL across three frozen backbones and two aggregators while holding downstream architectures constant within each paired comparison. Replacing conventional single-embedding bags with DI-MIL yields performance improvements in 67 out of 72 evaluation settings. Aggregate improvements are consistently positive across datasets, ranging from +1.10 to +3.21 percentage points. Fold-paired analyses in Appendix~\ref{app:paired_stats} further characterize the variability of these gains across matched folds.
The mean performance gains are most pronounced in liquid-based and \textit{Respiratory} cohorts, \textit{Serous Fluid} and \textit{Respiratory}, where rare diagnostic cells are diluted by abundant background. Decoupling instance granularity enables slide attention to focus on evidence-enriched sub-tile clusters. Across feature extractors, mean performance gains are highest under Smart-CCS (+3.64 points), followed by UniCAS (+1.59 points) and DINOv3 (+0.51 points). This trend indicates that DI-MIL leverages fine-grained representations more effectively when dense tokens capture domain-specific cytomorphological details. We further analyze this backbone-dependent behavior through quantitative and qualitative dense-token analyses in Appendix~\ref{app:cluster_visualization}. Furthermore, performance gains remain consistent between ABMIL (+1.86 points) and ACMIL (+1.88 points), demonstrating transferability across distinct attention aggregators.

\subsection{Comparison with Representative MIL Baselines}
\label{sec:second_result}
\vspace{-4pt}
\begin{table}[!ht]
\centering
\caption{
Comparison with existing MIL methods across four cytopathology datasets and three foundation-model backbones.
All methods within each backbone use identical features, data splits, and training protocols.
\textbf{DI-ACMIL} denotes ACMIL operating on our proposed decomposed instances.
Best results are shown in \textbf{bold}, and second-best results are \underline{underlined}.
}
\label{tab:sota_results}

\scriptsize
\setlength{\tabcolsep}{4pt}
\renewcommand{\arraystretch}{1.05}

\begin{tabular}{l|ccc|ccc|ccc|ccc}
\toprule
\multirow{2}{*}{Method}
& \multicolumn{3}{c|}{Cervical}
& \multicolumn{3}{c|}{Serous Fluid}
& \multicolumn{3}{c|}{Respiratory}
& \multicolumn{3}{c}{Oral2026} \\
\cmidrule(lr){2-4}
\cmidrule(lr){5-7}
\cmidrule(lr){8-10}
\cmidrule(lr){11-13}
& AUC & ACC & F1
& AUC & ACC & F1
& AUC & ACC & F1
& AUC & ACC & F1 \\
\midrule

\multicolumn{13}{c}{\textbf{DINOv3}} \\
\midrule

ABMIL
& 91.40 & 84.56 & 84.40
& 85.37 & 78.76 & 78.70
& 79.38 & 74.73 & 70.64
& 88.06 & 79.84 & 70.93 \\

ACMIL
& \underline{93.83}
& \underline{87.22}
& \underline{87.01}
& \underline{86.11}
& 79.13
& 79.11
& \underline{81.05}
& \underline{75.98}
& \underline{72.11}
& \underline{88.37}
& 80.77
& 71.45 \\

DTFD
& 92.29 & 86.00 & 85.66
& 85.13 & 78.06 & 77.99
& 75.77 & 71.57 & 66.90
& 78.98 & 81.50 & 58.29 \\

ASMIL
& 92.57 & 87.02 & 86.89
& 84.31 & 77.40 & 77.32
& 73.11 & 74.54 & 69.95
& 85.02 & 80.65 & 58.73 \\

DSMIL
& 90.00 & 83.84 & 83.59
& 83.06 & 76.52 & 76.46
& 74.42 & 72.43 & 68.16
& 85.60 & \underline{82.67} & \textbf{72.57} \\

TransMIL
& 91.32 & 83.44 & 83.21
& 85.03 & 77.40 & 77.35
& 75.79 & 71.47 & 66.50
& 79.52 & 72.53 & 63.87 \\

MambaMIL
& 90.94 & 86.71 & 86.48
& 86.00 & \underline{79.24} & \underline{79.19}
& 80.38 & 74.06 & 70.34
& 83.22 & \textbf{82.90} & 63.85 \\

HMIL
& 91.93 & 84.15 & 83.84
& 84.78 & 77.14 & 77.03
& 79.19 & 74.16 & 70.28
& 82.16 & 81.51 & 65.58 \\

PTCMIL
& 91.86 & 84.66 & 84.44
& 85.82 & 78.50 & 78.44
& 79.64 & 74.16 & 68.46
& 86.14 & 80.60 & 64.33 \\

\greyrow
DI-ACMIL (Ours)
& \textbf{93.86}
& \textbf{88.24}
& \textbf{88.06}
& \textbf{86.79}
& \textbf{80.09}
& \textbf{80.06}
& \textbf{82.01}
& \textbf{76.75}
& \textbf{72.67}
& \textbf{89.36}
& 79.91
& \underline{72.47} \\

\midrule
\multicolumn{13}{c}{\textbf{Smart-CCS}} \\
\midrule

ABMIL
& 94.38
& 87.42
& 87.28
& 84.22 & 76.52 & 76.49
& 79.36 & 74.16 & 70.04
& 83.91 & 83.18 & 67.18 \\

ACMIL
& 94.31 & 86.82 & 86.61
& \underline{85.04}
& \underline{77.51}
& \underline{77.43}
& \underline{81.21}
& \underline{76.17}
& \underline{71.82}
& \underline{88.20}
& 81.73
& 72.26 \\

DTFD
& 92.37 & 86.71 & 86.55
& 83.91 & 76.30 & 76.28
& 76.38 & 71.37 & 67.48
& 82.19 & 83.16 & 51.88 \\

ASMIL
& 91.20 & 86.61 & 86.48
& 78.77 & 74.35 & 74.32
& 72.91 & 73.39 & 68.93
& 78.26 & 81.79 & 65.19 \\

DSMIL
& 93.90 & 87.22 & 87.10
& 81.88 & 73.76 & 73.68
& 75.55 & 70.60 & 65.80
& 84.65 & 80.92 & 70.49 \\

TransMIL
& 93.30 & 87.12 & 86.95
& 83.17 & 75.45 & 75.36
& 75.85 & 70.99 & 66.04
& 86.54 & 76.92 & 68.15 \\

MambaMIL
& 92.92 & \underline{87.74} & \underline{87.65}
& 83.00 & 75.52 & 75.46
& 79.16 & 74.07 & 70.11
& 84.32 & 83.61 & 67.57 \\

HMIL
& 93.65 & 87.12 & 87.04
& 83.39 & 75.52 & 75.44
& 78.06 & 73.39 & 69.91
& 84.71 & 84.05 & 73.48 \\

PTCMIL
& \underline{94.61} & 87.02 & 86.82
& 84.87 & 76.59 & 76.46
& 79.13 & 73.39 & 69.00
& 87.79 & \underline{84.08} & \underline{73.75} \\

\greyrow
DI-ACMIL (Ours)
& \textbf{95.89}
& \textbf{89.98}
& \textbf{89.87}
& \textbf{88.33}
& \textbf{80.64}
& \textbf{80.60}
& \textbf{88.85}
& \textbf{82.32}
& \textbf{80.08}
& \textbf{89.13}
& \textbf{84.52}
& \textbf{75.69} \\

\midrule
\multicolumn{13}{c}{\textbf{UniCAS}} \\
\midrule

ABMIL
& 93.68 & 88.66 & 88.54
& 84.04
& \underline{76.85}
& \underline{76.81}
& 75.92 & 73.78
& \underline{69.89}
& 85.92
& \underline{85.30}
& 71.74 \\

ACMIL
& \underline{95.42}
& \underline{89.06}
& \underline{88.84}
& \underline{84.45}
& 76.52 & 76.48
& \underline{78.70}
& \underline{74.54}
& 69.63
& \underline{87.62}
& 84.82
& 71.63 \\

DTFD
& 92.66 & 86.30 & 86.09
& 82.76 & 74.97 & 74.93
& 70.90 & 68.30 & 63.07
& 85.06
& \textbf{86.10}
& \underline{72.42} \\

ASMIL
& 90.58 & 87.43 & 87.30
& 80.63 & 74.31 & 74.28
& 72.07 & 72.14 & 67.04
& 85.73 & 82.05 & 65.13 \\

DSMIL
& 92.17 & 86.20 & 85.98
& 80.31 & 73.17 & 73.08
& 71.99 & 71.09 & 64.88
& 82.39 & 83.39 & 69.75 \\

TransMIL
& 92.06 & 84.97 & 84.78
& 80.18 & 73.17 & 73.12
& 70.33 & 68.11 & 60.96
& 86.52 & 84.25 & 64.57 \\

MambaMIL
& 93.34 & 86.51 & 86.31
& 83.12 & 76.22 & 76.15
& 76.02 & 71.18 & 65.98
& 83.22 & 82.20 & 70.03 \\

HMIL
& 93.55 & 87.02 & 86.91
& 82.04 & 75.08 & 74.95
& 74.07 & 71.75 & 68.71
& 85.21 & 82.80 & 68.86 \\

PTCMIL
& 94.68 & 87.73 & 87.54
& 83.68 & 74.05 & 73.83
& 76.25 & 72.14 & 65.70
& 86.31 & 83.71 & 70.24 \\

\greyrow
DI-ACMIL (Ours)
& \textbf{95.79}
& \textbf{89.57}
& \textbf{89.39}
& \textbf{86.06}
& \textbf{78.43}
& \textbf{78.39}
& \textbf{80.73}
& \textbf{75.70}
& \textbf{71.41}
& \textbf{88.76}
& 84.60
& \textbf{74.38} \\

\bottomrule
\end{tabular}
\end{table}
\vspace{-8pt}

Decoupling instance construction proves competitive with complex architectural modifications to downstream aggregators. Table~\ref{tab:sota_results} compares DI-MIL paired with ACMIL against seven representative MIL baselines under identical backbone features and data splits. DI-MIL with ACMIL achieves the highest performance in \textbf{33 out of 36} backbone, dataset, and metric combinations, while obtaining the second-highest result in one additional setting (DINOv3 on \textit{Oral2026} macro-F1: 72.47 versus 72.57 for DSMIL). Detailed mean with standard deviation results are provided in Appendix \ref{app:detailed_results}.

Performance improvements on \textit{Cervical} are comparatively gradual across all feature backbones. Diagnostic cellular morphology in cervical smears is relatively prominent at the tile level, reducing representation dilution. Furthermore, Smart-CCS and UniCAS incorporate cervical cytopathology during pretraining, establishing strong baseline representations that limit additional gains from instance decomposition.

The strongest performance gains occur under Smart-CCS features, where DI-MIL with ACMIL achieves the highest mean performance across all $12$ dataset and metric settings. Minor exceptions occur in classification accuracy on \textit{Oral2026}: MambaMIL achieves $82.90$ ACC using DINOv3 compared to $79.91$ for DI-MIL with ACMIL, whereas DTFD achieves $86.10$ ACC using UniCAS compared to $84.60$ for DI-MIL with ACMIL. These results demonstrate that while DI-MIL does not lead every individual metric on smaller imbalanced cohorts, it maintains consistent cross-backbone competitiveness.


\subsection{Ablation Studies}
We use the ablation studies to separate three potential sources of improvement. First, we vary the number of decomposed instances \(k\) to assess granularity sensitivity. Second, with \(k\) fixed, we compare feature-based clustering against random and spatial grouping to isolate the effect of the grouping criterion. Lastly, we compare post-encoding decomposition with direct smaller tiling. The full ablation suite is evaluated on both \textit{Oral2026} and \textit{Respiratory}. 
\subsubsection{Number of decomposed instances}
\label{sec:ablation_k}

We first vary \(k\in\{1,2,3,4,5\}\) on \textit{Oral2026} and \textit{Respiratory} using Smart-CCS with ACMIL while holding all other settings fixed. As shown in Table~\ref{tab:ccs_acmil_k_ablation}, \(k=4\) provides the most balanced performance across the two datasets: it achieves the highest AUC and macro-F1 on \textit{Oral2026} and the highest ACC and macro-F1 on \textit{Respiratory}, while \(k=5\) yields the highest \textit{Oral2026} ACC and \textit{Respiratory} AUC. The non-monotonic trend indicates that simply increasing the number of instances is not sufficient. We therefore use \(k=4\) as a practical fixed default rather than claiming a universal optimum. Importantly, the $k=1$ baseline serves as a control that isolates the feature representation change from the multi-instance decomposition effect by collapsing dense tokens into a single instance. As shown in Appendix~\ref{app:k1_control}, $k=4$ consistently outperforms both the original [CLS] baseline and this $k=1$ control across all datasets.
\providecommand{\res}[2]{#1{\scriptsize\,(#2)}}
\providecommand{\bestres}[2]{\textbf{#1}{\scriptsize\,(#2)}}
\vspace{-4pt}
\begin{table}[!ht]
\centering
\caption{
Configuration sweep over the number of decomposed instances \(k\) on \textit{Oral2026} and \textit{Respiratory} using Smart-CCS and ACMIL.
All configurations use identical data splits and downstream training settings within each dataset and differ only in \(k\).
Results are reported as mean with standard deviation in parentheses.
}
\label{tab:ccs_acmil_k_ablation}
\scriptsize
\setlength{\tabcolsep}{5pt}
\renewcommand{\arraystretch}{1.08}
\begin{tabular}{c|ccc|ccc}
\toprule
\multirow{2}{*}{$k$}
& \multicolumn{3}{c|}{Oral2026}
& \multicolumn{3}{c}{Respiratory} \\
\cmidrule(lr){2-4}
\cmidrule(lr){5-7}
& AUC & ACC & F1
& AUC & ACC & F1 \\
\midrule
1
& \res{88.68}{3.93}
& \res{83.98}{3.03}
& \res{75.19}{4.25}
& \res{84.08}{2.16}
& \res{78.10}{2.97}
& \res{74.51}{3.34} \\
2
& \res{88.86}{4.22}
& \res{84.28}{4.43}
& \res{75.17}{6.72}
& \res{84.94}{2.85}
& \res{77.43}{3.34}
& \res{74.01}{4.01} \\
3
& \res{88.79}{5.35}
& \res{84.00}{6.63}
& \res{75.57}{9.55}
& \res{87.00}{2.71}
& \res{79.25}{3.50}
& \res{76.08}{4.72} \\
4
& \bestres{89.13}{4.86}
& \res{84.52}{5.25}
& \bestres{75.69}{8.27}
& \res{88.85}{2.07}
& \bestres{82.32}{3.10}
& \bestres{80.08}{3.48} \\
5
& \res{87.68}{6.23}
& \bestres{85.06}{5.35}
& \res{74.81}{11.59}
& \bestres{89.32}{2.35}
& \res{82.03}{4.64}
& \res{79.18}{6.05} \\
\bottomrule
\end{tabular}
\end{table}
\vspace{-12pt}

\subsubsection{Alternative dense-token grouping strategies}
\label{sec:ablation_on_other_dense_grouping_strategy}
Having fixed \(k=4\), we next ask whether the criterion used to form the four instances matters. Random grouping removes both semantic and spatial structure, whereas spatial \(2\times2\) grouping preserves local position but ignores feature similarity. As shown in Table~\ref{tab:other_strategy_vs_di}, feature-based DI achieves the highest mean AUC, ACC, and macro-F1 on both \textit{Oral2026} and \textit{Respiratory}. The consistent advantage over both controls indicates that the gain is not explained merely by expanding each tile into four arbitrary instances. Detailed formulations of the two controls are provided in Appendix~\ref{app:alternative_grouping}.



\vspace{-8pt}
\begin{table}[!ht]
\centering
\caption{
Comparison of alternative dense-token grouping strategies on \textit{Oral2026} and \textit{Respiratory} using Smart-CCS and ACMIL.
All strategies construct \(k=4\) instances from the same dense-token representations and use identical downstream training settings.
Results are reported as mean with standard deviation in parentheses.
}
\label{tab:other_strategy_vs_di}
\scriptsize
\setlength{\tabcolsep}{4.5pt}
\renewcommand{\arraystretch}{1.08}
\begin{tabular}{l|ccc|ccc}
\toprule
\multirow{2}{*}{Strategy}
& \multicolumn{3}{c|}{Oral2026}
& \multicolumn{3}{c}{Respiratory} \\
\cmidrule(lr){2-4}
\cmidrule(lr){5-7}
& AUC & ACC & F1
& AUC & ACC & F1 \\
\midrule
Random token grouping ($k=4$)
& \res{88.07}{4.41}
& \res{83.75}{3.50}
& \res{74.79}{5.68}
& \res{83.28}{2.42}
& \res{76.95}{2.78}
& \res{73.60}{3.85} \\
Spatial $2 \times 2$ grouping
& \res{86.83}{5.56}
& \res{83.38}{4.88}
& \res{73.95}{7.29}
& \res{86.11}{2.22}
& \res{80.11}{1.93}
& \res{77.15}{2.61} \\
DI ($k=4$)
& \bestres{89.13}{4.86}
& \bestres{84.52}{5.25}
& \bestres{75.69}{8.27}
& \bestres{88.85}{2.07}
& \bestres{82.32}{3.10}
& \bestres{80.08}{3.48} \\
\bottomrule
\end{tabular}
\end{table}
\vspace{-8pt}

\subsubsection{Decomposed Instances versus Direct Fine-Grid Tiling}
\label{sec:ablation_strategy}

The previous comparison shows that post-encoding grouping matters even when the number of instances is fixed. We next test whether comparable gains can instead be obtained by increasing instance granularity before encoding through smaller source tiles. Table~\ref{tab:tile_size_vs_di} compares DI-MIL with progressively finer tiling on \textit{Oral2026} and \textit{Respiratory}. On \textit{Oral2026}, smaller tiling produces non-monotonic changes across tile sizes; on \textit{Respiratory}, smaller tiles progressively improve performance over the native 1200-pixel baseline. Nevertheless, DI-MIL achieves the highest mean AUC, ACC, and macro-F1 on both datasets while retaining the original source-tile encoding context. Thus, its gains cannot be attributed solely to a larger number of MIL instances.

\begin{figure}[!h]
    \centering
    \includegraphics[width=0.85\textwidth]{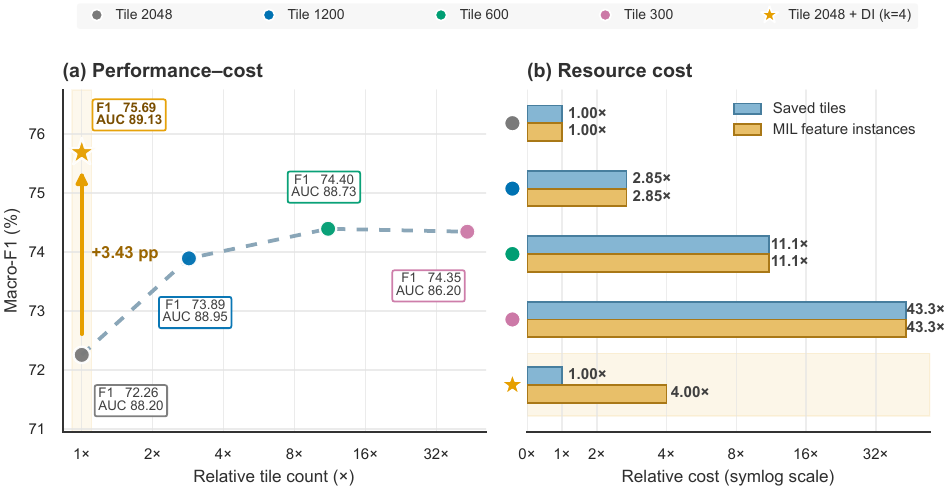}
    \caption{\textbf{Performance and computational cost comparison between source tile size reduction and DI-MIL on \textit{Oral2026} using Smart-CCS with ACMIL.} Left: AUC and macro-F1 versus the relative number of  image tiles. Right: Relative tile count versus MIL feature-instance count.}
    \label{fig:tile_cost}
\end{figure}

Additionally, we analyze the storage cost which serves as a major practical bottleneck in cytopathology pipelines due to high-resolution tile storage and slide-level I/O. As illustrated for \textit{Oral2026} in Figure~\ref{fig:tile_cost}, direct smaller tiling increases the number of extracted source images by up to \(43.3\times\), whereas DI-MIL preserves the original \(1\times\) image extraction and encoder cost while extending bag length from $N$ to $4N$. Absolute feature-storage measurements and cost analysis by computational time are further provided in Appendix \ref{app:storage_cost} \ref{app:cost_analysis_by_time}.

\begin{table}[!ht]
\centering
\caption{
Comparison of different tile-size strategies with our decomposed-instance (DI) strategy on \textit{Oral2026} and \textit{Respiratory}.
Tile \(n\) stands for using pixel tile size \(n\) without DI.
All methods use Smart-CCS features, ACMIL, and the same downstream training protocol;
DI uses the main-experiment setting \(k=4\) with the dataset-specific source tile size (2048 on \textit{Oral2026} and 1200 on \textit{Respiratory}).
Results are reported as mean with standard deviation in parentheses.
}
\label{tab:tile_size_vs_di}
\scriptsize
\setlength{\tabcolsep}{4.5pt}
\renewcommand{\arraystretch}{1.08}
\begin{tabular}{l|ccc|ccc}
\toprule
\multirow{2}{*}{Strategy}
& \multicolumn{3}{c|}{Oral2026}
& \multicolumn{3}{c}{Respiratory} \\
\cmidrule(lr){2-4}
\cmidrule(lr){5-7}
& AUC & ACC & F1
& AUC & ACC & F1 \\
\midrule
Tile 2048 w/o.\ DI
& \res{88.20}{5.44}
& \res{81.73}{6.14}
& \res{72.26}{4.25}
& -- & -- & -- \\
Tile 1200 w/o.\ DI
& \res{88.95}{5.06}
& \res{84.26}{4.78}
& \res{73.89}{4.73}
& \res{81.21}{3.45}
& \res{76.17}{3.19}
& \res{71.83}{4.47} \\
Tile 600 w/o.\ DI
& \res{88.73}{3.83}
& \res{84.24}{5.83}
& \res{74.40}{4.75}
& \res{83.81}{3.92}
& \res{78.10}{3.39}
& \res{74.77}{3.22} \\
Tile 300 w/o.\ DI
& \res{86.20}{3.34}
& \res{84.10}{4.88}
& \res{74.35}{4.84}
& \res{85.49}{2.11}
& \res{78.29}{2.93}
& \res{75.07}{3.78} \\
DI ($k=4$)
& \bestres{89.13}{4.86}
& \bestres{84.52}{5.25}
& \bestres{75.69}{8.27}
& \bestres{88.85}{2.07}
& \bestres{82.32}{3.10}
& \bestres{80.08}{3.48} \\
\bottomrule
\end{tabular}
\end{table}
\vspace{-8pt}
\vspace{-4pt}

\vspace{2pt}
\section{Discussion and Limitations}
\label{sec:discussion}

Conventional WSI MIL operationally treats each encoded tile as an indivisible unit of slide-level aggregation. This formulation is particularly restrictive in cytopathology, where a single microscopic field may contain sparse diagnostic cells intermixed with abundant benign and non-diagnostic content. DI-MIL addresses this fundamental domain mismatch by decoupling visual encoding context from slide aggregation granularity, enabling attention-based aggregators such as ABMIL and ACMIL to assign independent importance weights to localized, discriminative cellular patterns. Qualitative inspection of high-attention instances is consistent with this intended granularity effect; examples are provided in Appendix~\ref{app:attention_visualization}. 

On the other hand, DI-MIL has two main limitations. First, its fixed \(k\) and non-differentiable clustering do not adapt to variations in tile cellularity or spatial heterogeneity. Second, it treats decomposed instances as an order-agnostic set and therefore does not explicitly model spatial relationships among intra-tile clusters. Potential extensions include adaptive or differentiable decomposition, spatially aware grouping, and task-guided interactions among decomposed instances.
\section{Conclusion}
\label{sec:conclusion}

We revisited the conventional paradigm in weakly supervised cytopathology that couples tile-level encoding context to MIL aggregation granularity. By decomposing dense tokens from foundation models within each tile into independently weighted instances post-encoding, DI-MIL mitigates the intra-tile evidence dilution that can arise when heterogeneous cytopathology tiles are compressed into single global representations. Extensive evaluations across diverse cytopathology cohorts, foundation backbones, and aggregators demonstrate that DI-MIL consistently outperforms both single-embedding baselines and direct fine-grid tiling. These results demonstrate that the tile-level instance is not always the most effective choice for cytopathology MIL and establish instance construction as a meaningful direction for future study.

\subsection*{AI use statement}
Generative AI tools were used during preparation of this manuscript to assist with literature search and organization, consistency checking and interpretation of already-computed experimental outputs, and drafting and editing of the paper. The experimental measurements reported in the tables and figures were supplied from the authors' experimental pipeline rather than generated by the language model used for manuscript preparation. The authors reviewed the AI-assisted prose, recomputed the aggregate numerical summaries used in the text from the supplied results, and checked cited claims against the referenced literature. The authors take responsibility for the final content of the work, including text and claims produced with generative-AI assistance.

\subsection*{ETHICS STATEMENT}
The private cytopathology cohorts were retrospectively collected and processed under the applicable institutional ethics and data-governance requirements. The study protocol was reviewed and approved by the relevant institutional ethics committee, and the requirement for informed consent was waived because of the retrospective nature of the study. All WSIs and associated labels were de-identified before use. The study involved retrospective model-development experiments only and did not include any prospective clinical intervention. The public \textit{Oral2026} dataset was used in accordance with its released terms. The proposed models are intended for research use and have not been validated for independent clinical deployment.

\subsection*{Reproducibility statement}
Section~\ref{sec:method} specifies the instance-construction operation and its integration with the evaluated attention-based MIL backends; Section~\ref{sec:setup} defines the controlled comparison axes; and Appendix~\ref{app:setup_details} reports cohort sizes, tile sizes, cross-validation protocols, split seeds, core training hyperparameters, and the default decomposition configuration. The appendix also provides an algorithmic summary, fold-wise statistics for representative baselines, and the cost accounting used in the tile-size ablation. All paired comparisons use the same underlying features, data splits, and downstream training protocol except for the explicitly varied instance-construction factor. Exact foundation-model checkpoint identifiers, feature-extraction preprocessing conventions, and implementation-level clustering details will accompany the anonymized code/configuration release.

\bibliography{iclr2027_conference}
\bibliographystyle{iclr2027_conference}

\clearpage
\appendix
\section{Appendix}
\label{sec:appendix}
\raggedbottom

\subsection{Datasets, Splits, and Training Protocol}
\label{app:setup_details}
Table~\ref{tab:dataset_details} summarizes the cohort and evaluation settings. \textit{Cervical}, \textit{Serous Fluid}, and \textit{Respiratory} are private cytopathology cohorts scanned at \(0.5\,\mu\text{m/pixel}\). We extract non-overlapping \(1200\times1200\)-pixel tiles using a stride of 1200 pixels after Otsu-based background filtering.

The same folds are used by all methods, and no tiles from the same WSI appear in different folds. Reported values are the mean and standard deviation across the corresponding cross-validation folds.

The token grid is determined by the backbone input resolution after preprocessing and is distinct from the native WSI source-tile size reported in Table~\ref{tab:dataset_details}.

\begin{table}[!htbp]
\centering
\caption{Dataset and evaluation settings used in the experiments. Positive rate denotes the positive/cancerous class fraction. Avg. tiles / slide denotes the mean number of retained source tiles per WSI after background filtering.}
\label{tab:dataset_details}
\scriptsize
\setlength{\tabcolsep}{4.0pt}
\renewcommand{\arraystretch}{1.08}
\begin{tabular}{lccccccc}
\toprule
Dataset & Specimen & Slides & Task & Positive rate & Tile size & CV & Avg. tiles / slide\\
\midrule
Cervical & Cervical & 978 & Negative vs. Positive & 53.3\% & 1200 px & 10-fold & 1940 \\
Serous Fluid & Serous fluid & 2,717 & Negative vs. Positive & 51.5\% & 1200 px & 10-fold & 1572 \\
Respiratory & Respiratory & 1,041 & Negative vs. Positive & 66.3\% & 1200 px & 10-fold & 1141 \\
Oral2026 & Oral & 363 & Non-cancerous vs. Cancerous & 17.6\% & 2048 px & 5-fold & 2579 \\
\bottomrule
\end{tabular}
\end{table}

\begin{table}[!htbp]
\centering
\caption{Details of the backbones used in the experiments.}
\label{tab:backbone_details}
\scriptsize
\setlength{\tabcolsep}{18.0pt}
\renewcommand{\arraystretch}{1.08}
\begin{tabular}{lccc}
\toprule
Backbone & Encoder patch size (px) & Token grid & Dense tokens / encoded tile \\
\midrule
Smart-CCS & 14 & 16 $\times$ 16 & 256 \\
DINOv3 (vit-l16) & 16 & 14 $\times$ 14 & 196 \\
UniCAS & 16 & 14 $\times$ 14 & 196 \\
\bottomrule
\end{tabular}
\end{table}
These statistics also clarify the scale of the instance-construction step. A conventional bag contains \(N\) tile embeddings, whereas DI-MIL with \(k=4\) contains \(4N\) instances. Directly exposing all dense patch tokens to the MIL aggregator would instead produce \(196N\) or \(256N\) instances depending on the backbone, motivating the intermediate clustering step.

Across datasets, backbones, and the paired MIL experiments, training is standardized wherever the backend permits. Table~\ref{tab:training_details} lists the common settings and the backend-specific choices. \textit{Oral2026} uses class-weighted cross-entropy because of its smaller and more imbalanced cohort.

\begin{table}[H]
\centering
\caption{Core training and decomposition settings. Backend-specific optimizer/objective choices follow the corresponding official MIL formulation.}
\label{tab:training_details}
\scriptsize
\setlength{\tabcolsep}{5pt}
\renewcommand{\arraystretch}{1.08}
\begin{tabular}{p{0.28\textwidth}p{0.66\textwidth}}
\toprule
Setting & Value \\
\midrule
Learning rate & $2\times 10^{-4}$ \\
Training seed & 1024 \\
Optimizer & Adam (ABMIL); AdamW (ACMIL) \\
Objective & CE (ABMIL); sub-token CE + slide CE + attention-diversity loss (ACMIL) \\
Default decomposition & $k=4$; 8 clustering iterations \\
Early stopping / maximum epochs & Enabled / 100 \\
Feature dimension / classes & 1024 / 2 \\
Input normalization & LayerNorm enabled \\
Batch size & 1 WSI (one bag) \\
Weight decay / dropout & $10^{-5}$ / 0.25 \\
Metrics & AUC, ACC, macro-F1 \\
\bottomrule
\end{tabular}
\end{table}

\clearpage

\subsection{Analysis of Fine-Grained Dense-Token Representations}
\label{app:cluster_visualization}
The main experiments show that DI-MIL yields larger improvements with the cytopathology-specific Smart-CCS and UniCAS backbones than with the general-purpose DINOv3 backbone. To further examine this backbone-dependent behavior, we analyze whether their dense-token representations differ in their ability to preserve localized cellular evidence, using both quantitative evaluation and qualitative visualization.

\begin{figure}[ht] 
    \centering 
    \includegraphics[width=0.720\linewidth] {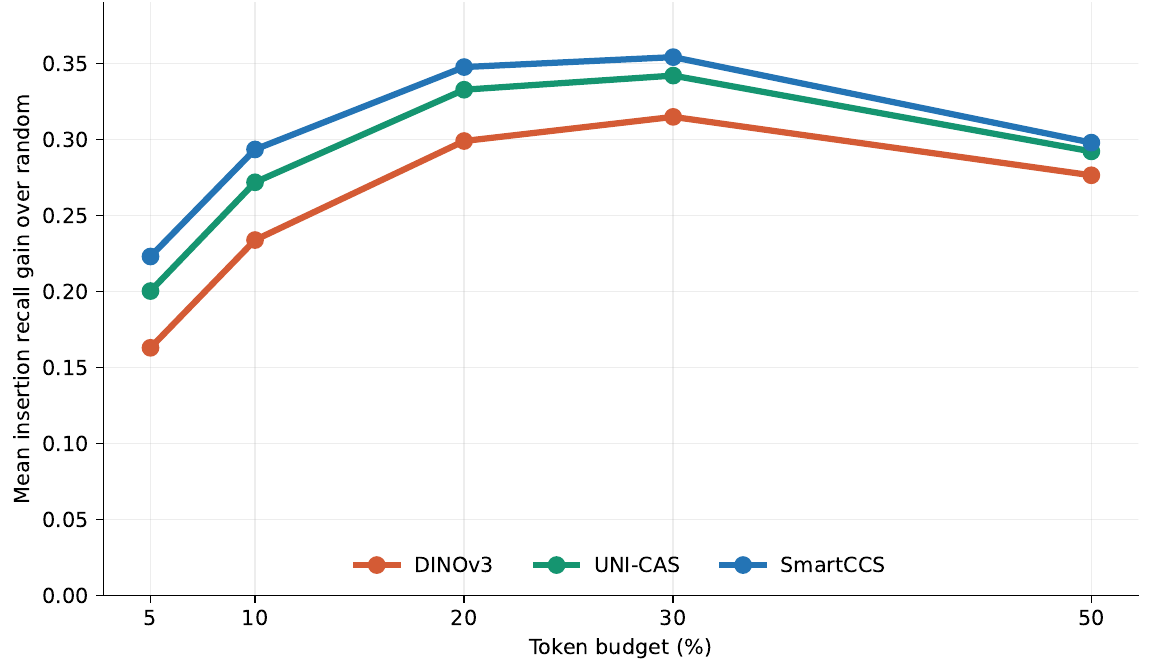} 
    \caption{Mean random-corrected insertion-recall gain across 18 public cell-level datasets as a function of retained token budget. Higher values indicate that annotated cellular evidence is recovered more readily from the frozen dense-token representation.} \label{fig:insertion_curve} 
\end{figure}

\paragraph{Quantitative analysis.} We evaluate the frozen dense-token representations on 18 public datasets with cell-level bounding-box annotations or bounding boxes derived from available instance masks. For each annotated object, we extract a $224\times224$ crop centered on its ground-truth bounding box and obtain the corresponding dense patch-token embeddings. A token is regarded as positive evidence if the center of its spatial support lies within the annotated bounding box. Using validation data only, we compute positive- and negative-evidence prototypes, denoted by $\boldsymbol{\mu}_{+}$ and $\boldsymbol{\mu}_{-}$. For each test token $\mathbf{z}_i$, we define an evidence score as \begin{equation} s_i = \sigma\left[ \cos(\mathbf{z}_i,\boldsymbol{\mu}_{+}) - \cos(\mathbf{z}_i,\boldsymbol{\mu}_{-}) \right], \end{equation} where $\sigma(\cdot)$ denotes the sigmoid function. Test tokens are ranked by $s_i$, and we measure how rapidly tokens corresponding to annotated cellular evidence are recovered as progressively larger fractions of the ranked tokens are included. We summarize this behavior using the random-corrected insertion AUC gain \begin{equation} G = \operatorname{AUC}(I) - \operatorname{AUC}(I_{\mathrm{rnd}}), \end{equation} where $I$ denotes the insertion-recall curve and $I_{\mathrm{rnd}}$ denotes the corresponding curve under a deterministic random token ordering. A larger $G$ indicates that annotated cellular evidence is more readily identifiable from the frozen dense-token representation. Across the 18 datasets, Smart-CCS achieves the highest equal-weight mean insertion gain ($G=0.2254$), followed by UniCAS ($0.2173$) and DINOv3 ($0.1996$). Smart-CCS and UniCAS outperform DINOv3 on 14/18 and 15/18 datasets, respectively, with one-sided paired Wilcoxon $p$-values of $0.0014$ and $0.00042$. As shown in Figure~\ref{fig:insertion_curve}, both cytopathology-specific backbones also achieve higher mean insertion gains than DINOv3 across token budgets from 5\% to 50\%. Detailed dataset-level results are provided in Table~\ref{tab:detailed_dense_token_results}. 

\begin{table}[H]
\centering
\caption{
Dataset-level random-corrected insertion AUC gain $G$ for the three frozen backbones.
Higher values indicate that annotated cellular evidence is recovered more readily from the dense-token representation.
$S-D$ and $U-D$ denote the paired differences of Smart-CCS and UniCAS relative to DINOv3, respectively.
}
\label{tab:detailed_dense_token_results}

\small
\setlength{\tabcolsep}{3.0pt}
\renewcommand{\arraystretch}{1.05}

\begin{tabular}{@{}lcccccc@{}}
\toprule
Dataset
& Smart-CCS
& UniCAS
& DINOv3
& $S-D$
& $U-D$
& Test objects \\
\midrule

APACS23~\citep{APACS23}
& 0.1023 & \textbf{0.1045} & 0.0983
& +0.0040 & +0.0062 & 8,671 \\

APC~\citep{APCData}
& \textbf{0.2263} & 0.2112 & 0.1569
& +0.0694 & +0.0542 & 281 \\

Ascites2020~\citep{Ascites2020}
& \textbf{0.4293} & 0.4213 & 0.3499
& +0.0794 & +0.0714 & 581 \\

BCCD~\citep{BCCD}
& 0.1279 & 0.1279 & \textbf{0.1286}
& -0.0007 & -0.0007 & 71 \\

BTTFA~\citep{BTTFA}
& \textbf{0.4551} & 0.4015 & 0.3914
& +0.0637 & +0.0101 & 978 \\

CBC~\citep{CBC}
& 0.0532 & 0.0485 & \textbf{0.0548}
& -0.0016 & -0.0063 & 61 \\

CellDet~\citep{CellDet}
& 0.1769 & \textbf{0.1816} & 0.1550
& +0.0219 & +0.0266 & 3,104 \\

CISD~\citep{CISD}
& \textbf{0.0807} & 0.0682 & 0.0636
& +0.0171 & +0.0046 & 921 \\

CNSeg~\citep{CNSeg}
& \textbf{0.4028} & 0.3752 & 0.3507
& +0.0521 & +0.0246 & 10,649 \\

CPS~\citep{jiang2023donet}
& 0.2667 & \textbf{0.2815} & 0.2241
& +0.0426 & +0.0573 & 1,209 \\

Herlev~\citep{Herlev}
& 0.1298 & \textbf{0.1325} & 0.1312
& -0.0014 & +0.0014 & 399 \\

Jiangxi~\citep{Jiangxi}
& 0.3180 & 0.3267 & \textbf{0.3270}
& -0.0091 & -0.0004 & 134 \\

NMCD~\citep{NMCD}
& \textbf{0.2424} & 0.2054 & 0.1577
& +0.0847 & +0.0477 & 1,057 \\

Oral2021~\citep{Oral2021}
& \textbf{0.3712} & 0.3587 & 0.3559
& +0.0153 & +0.0028 & 253 \\

Raabin-WBC~\citep{Raabin_WBC}
& \textbf{0.0917} & 0.0891 & 0.0827
& +0.0090 & +0.0064 & 461 \\

SegPC~\citep{SegPC}
& 0.0248 & \textbf{0.0249} & 0.0243
& +0.0005 & +0.0006 & 948 \\

TXL-PBC~\citep{TXL_PBC}
& 0.1389 & \textbf{0.1440} & 0.1383
& +0.0006 & +0.0057 & 155 \\

UFSC-OCPap~\citep{UFSC_OCPap}
& \textbf{0.4187} & 0.4084 & 0.4027
& +0.0160 & +0.0057 & 5,081 \\

\midrule

Equal-weight mean
& \textbf{0.2254}
& 0.2173
& 0.1996
& +0.0258
& +0.0177
& 35,014 \\

\bottomrule
\end{tabular}
\end{table}

\paragraph{Qualitative analysis.}
For qualitative visualization, the dense token feature maps are upsampled to the input-image resolution using AnyUp \citep{wimmer2026anyup}. This upsampling is used only for qualitative visualization and is not involved in token clustering, DI construction, or downstream MIL training.

\begin{figure}[ht]
    \centering
    \includegraphics[width=0.99\linewidth]{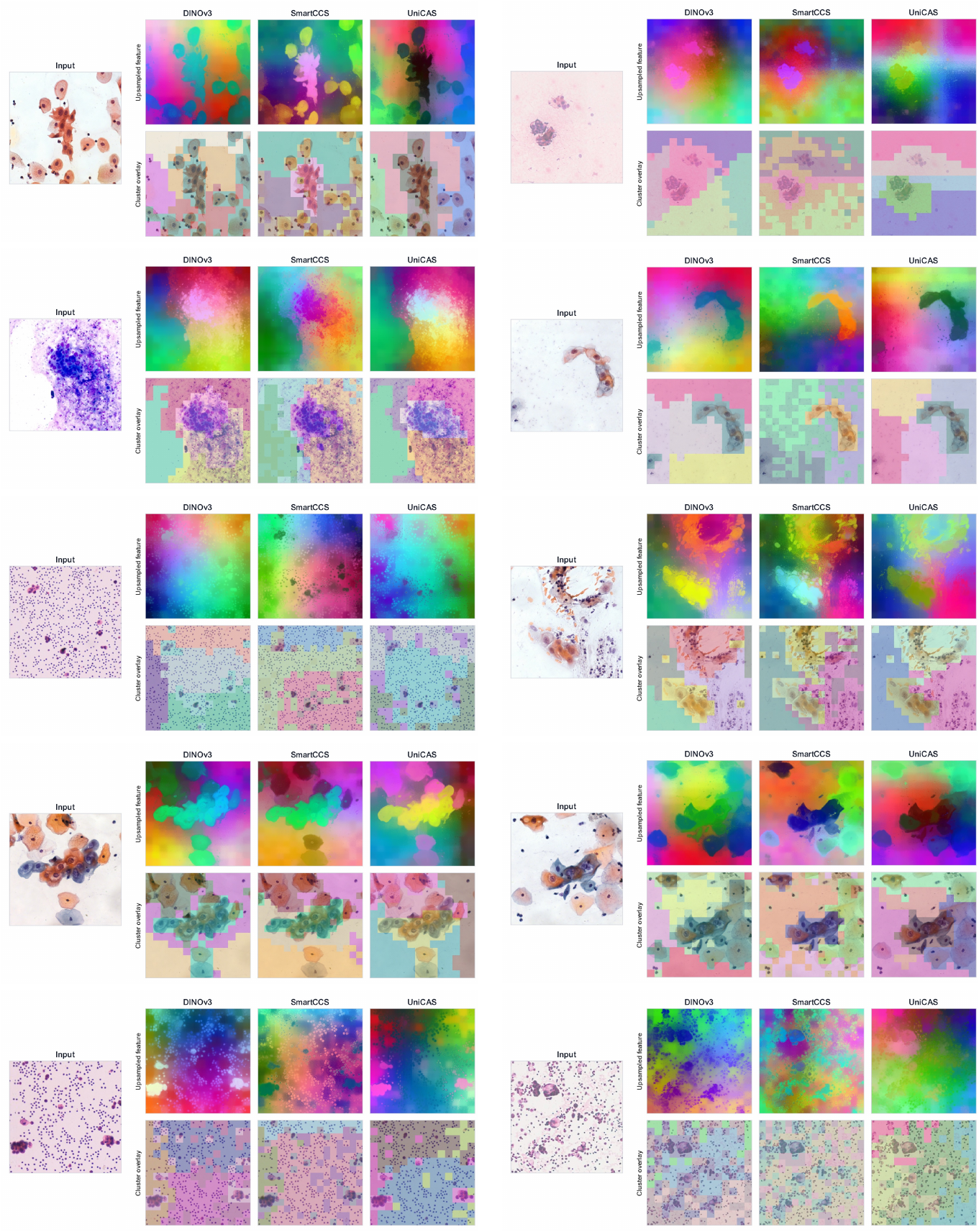}
    \caption{Visualization of dense token representations and decomposed instances. Patch tokens within the same color group form one DI. The color difference is positively correlated to feature vector distance in the encoder feature space.}
    \label{fig:app_cluster_visualization}
\end{figure}

Figure~\ref{fig:app_cluster_visualization} provides qualitative examples of the dense token representations from the three evaluated backbones and the corresponding intra-tile clustering used to construct decomposed instances. The examples suggest that the cytopathology-specific foundation models, Smart-CCS and UniCAS, exhibit more distinct dense-token representations of abnormal cytomorphological patterns than DINOv3.

\clearpage
\subsection{Alternative Dense-Token Grouping Strategies}
\label{app:alternative_grouping}

To isolate the effect of feature-based token grouping in DI-MIL, we compare the proposed spherical clustering with two alternative strategies that construct the same number of instances from the same dense patch-token representations. Specifically, we consider \emph{random grouping} and \emph{spatial $2\times2$ grouping}. Both alternatives operate independently within each source tile and use $k=4$, matching the default DI-MIL configuration. They differ from DI-MIL only in how patch tokens are assigned to groups; the subsequent instance pooling and slide-level MIL aggregation remain unchanged.

Let the $M$ patch-token representations extracted from tile $p_i$ be
$\{\vh_{ij}\}_{j=1}^{M}$, following the notation in Section~\ref{sec:method}.

\paragraph{Random grouping.}
Random grouping removes both feature similarity and spatial structure from the decomposition procedure. For each tile $p_i$, we randomly permute the token indices $\{1,\ldots,M\}$ and partition the resulting sequence into $k=4$ disjoint groups of approximately equal size,
\begin{equation}
    \{\mathcal{R}_{i1},\ldots,\mathcal{R}_{ik}\}
    =
    \operatorname{Split}_{k}\!\left(
        \operatorname{Permute}(\{1,\ldots,M\})
    \right),
    \qquad
    \bigcup_{\ell=1}^{k}\mathcal{R}_{i\ell}
    =
    \{1,\ldots,M\}.
    \label{eq:random-grouping}
\end{equation}
When $M$ is not exactly divisible by $k$, group sizes differ by at most one token. Each random group is then converted into one MIL instance using the same pooling operation as DI-MIL,
\begin{equation}
    \vd_{i,\mathrm{rand}}^{\ell}
    =
    \operatorname{LN}\!\left(
        \frac{1}{|\mathcal{R}_{i\ell}|}
        \sum_{j\in\mathcal{R}_{i\ell}}
        \vh_{ij}
    \right),
    \qquad
    \ell=1,\ldots,4.
    \label{eq:random-pooling}
\end{equation}
Thus, each source tile still contributes four independently weighted MIL instances, but membership in an instance is unrelated to either token semantics or spatial location. This baseline tests whether the performance gain can be explained simply by increasing the number of instances constructed from dense tokens.

\paragraph{Spatial $2\times2$ grouping.}
Spatial grouping instead preserves the original spatial organization of the transformer token grid while ignoring feature similarity. Suppose the $M$ patch tokens of tile $p_i$ form an $H\times W$ spatial grid, with token $j$ located at grid coordinate $(r_j,c_j)$. We divide this grid into four non-overlapping spatial quadrants according to the horizontal and vertical midpoints,
\begin{equation}
    \mathcal{S}_{i}^{ab}
    =
    \left\{
        j
        \;\middle|\;
        \mathbb{I}\!\left(r_j \geq H/2\right)=a,\,
        \mathbb{I}\!\left(c_j \geq W/2\right)=b
    \right\},
    \qquad
    a,b\in\{0,1\},
    \label{eq:spatial-grouping}
\end{equation}
yielding the four groups corresponding to the top-left, top-right, bottom-left, and bottom-right regions of the token grid. Each spatial group is pooled identically to DI-MIL,
\begin{equation}
    \vd_{i,\mathrm{sp}}^{ab}
    =
    \operatorname{LN}\!\left(
        \frac{1}{|\mathcal{S}_{i}^{ab}|}
        \sum_{j\in\mathcal{S}_{i}^{ab}}
        \vh_{ij}
    \right),
    \qquad
    a,b\in\{0,1\}.
    \label{eq:spatial-pooling}
\end{equation}

Spatial $2\times2$ grouping therefore also produces exactly four instances per tile, but constrains each instance to a contiguous spatial region. In contrast, DI-MIL groups tokens according to similarity in the encoder representation space and does not require grouped tokens to be spatially adjacent.

For all three strategies, the resulting instances retain the original feature dimensionality and are assembled across tiles into a slide-level bag of size $4N$. Consequently, the encoder inputs, dense-token representations, number of MIL instances, pooling operation, and downstream aggregator are held fixed. The comparison in Section~\ref{sec:ablation_on_other_dense_grouping_strategy} therefore primarily evaluates the effect of the token-grouping criterion itself: random assignment, fixed spatial partitioning, or feature-based clustering.

\subsection{Controlling for Dense-Token Mean Pooling (\(k=1\))}
\label{app:k1_control}
The main Tile--DI comparison changes both the representation exposed to the MIL model and the number of independently weighted instances: the Tile baseline uses the encoder [CLS] embedding, whereas DI is constructed from dense patch-token representations. We therefore use \(k=1\) as a representation control. With \(k=1\), all dense tokens within a tile are mean-pooled into a single instance, so the bag still contains exactly one instance per source tile. Across all four datasets using Smart-CCS with ACMIL, \(k=4\) outperforms both the [CLS] baseline and the \(k=1\) control on all 12 dataset--metric comparisons. The \(k=1\) representation is not uniformly better than [CLS], indicating that the main improvement cannot be explained solely by replacing the global token with dense-token mean pooling.

\vspace{-8pt}
\begin{table}[!ht]
\centering
\caption{
Comparison of the conventional tile-level [CLS] representation, dense-token mean pooling (\(k=1\)), and the default decomposed-instance setting (\(k=4\)) across all four datasets. All variants use Smart-CCS, ACMIL, identical dataset-native source tiles, and the same downstream training protocol.
}
\label{tab:k1_vs_cls_vs_di}
\scriptsize
\setlength{\tabcolsep}{4.5pt}
\renewcommand{\arraystretch}{1.08}

\begin{minipage}[t]{0.48\linewidth}
\centering
\begin{tabular}{lccc}
\toprule
\multicolumn{4}{c}{Oral2026} \\
\cmidrule(lr){1-4}
Method & AUC & ACC & F1 \\
\midrule
{[CLS]}
& \res{88.20}{5.44}
& \res{81.73}{6.14}
& \res{72.26}{4.25} \\

DI ($k=1$)
& \res{88.68}{3.93}
& \res{83.98}{3.03}
& \res{75.19}{4.25} \\

DI ($k=4$)
& \bestres{89.13}{4.86}
& \bestres{84.52}{5.25}
& \bestres{75.69}{8.27} \\
\bottomrule
\end{tabular}
\end{minipage}
\hfill
\begin{minipage}[t]{0.48\linewidth}
\centering
\begin{tabular}{lccc}
\toprule
\multicolumn{4}{c}{Cervical} \\
\cmidrule(lr){1-4}
Method & AUC & ACC & F1 \\
\midrule
{[CLS]}
& \res{94.31}{1.90}
& \res{86.82}{3.36}
& \res{86.61}{3.48} \\

DI ($k=1$)
& \res{94.66}{1.46}
& \res{88.65}{2.34}
& \res{88.46}{2.46} \\

DI ($k=4$)
& \bestres{95.89}{1.21}
& \bestres{89.98}{3.36}
& \bestres{89.87}{3.35} \\
\bottomrule
\end{tabular}
\end{minipage}

\vspace{0.8em}

\begin{minipage}[t]{0.48\linewidth}
\centering
\begin{tabular}{lccc}
\toprule
\multicolumn{4}{c}{Serous Fluid} \\
\cmidrule(lr){1-4}
Method & AUC & ACC & F1 \\
\midrule
{[CLS]}
& \res{85.04}{2.82}
& \res{77.51}{3.20}
& \res{77.43}{3.25} \\

DI ($k=1$)
& \res{84.16}{2.63}
& \res{76.44}{2.24}
& \res{76.41}{2.25} \\

DI ($k=4$)
& \bestres{88.33}{2.09}
& \bestres{80.64}{2.62}
& \bestres{80.60}{2.63} \\
\bottomrule
\end{tabular}
\end{minipage}
\hfill
\begin{minipage}[t]{0.48\linewidth}
\centering
\begin{tabular}{lccc}
\toprule
\multicolumn{4}{c}{Respiratory} \\
\cmidrule(lr){1-4}
Method & AUC & ACC & F1 \\
\midrule
{[CLS]}
& \res{81.21}{3.45}
& \res{76.17}{3.19}
& \res{71.83}{4.47} \\

DI ($k=1$)
& \res{84.08}{2.16}
& \res{78.10}{2.97}
& \res{74.51}{3.34} \\

DI ($k=4$)
& \bestres{88.85}{2.07}
& \bestres{82.32}{3.10}
& \bestres{80.08}{3.48} \\
\bottomrule
\end{tabular}
\end{minipage}

\end{table}
\vspace{-8pt}

\subsection{Qualitative Analysis of High-Attention Instances}
\label{app:attention_visualization}

\begin{figure}[H]
    \centering
    \includegraphics[width=0.90\textwidth]
    {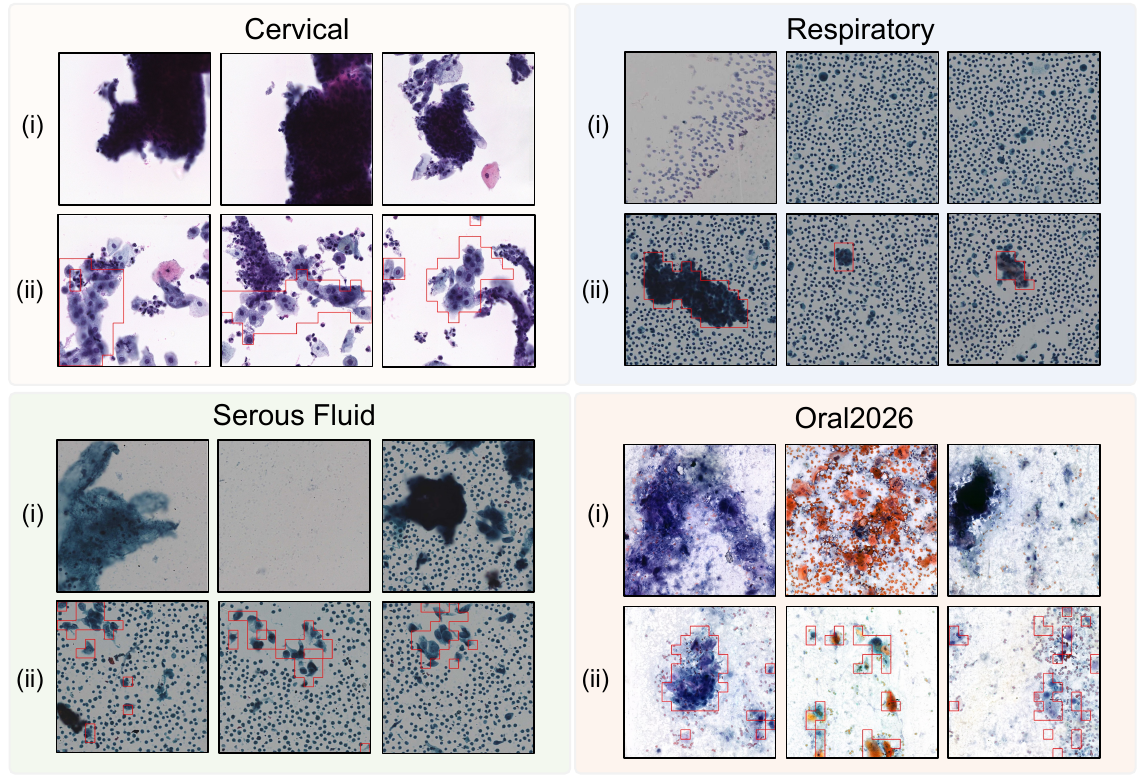}
    \caption{
    Qualitative comparison of high-attention instances across the four datasets.
    Row (i) shows conventional tile instances, whereas row (ii) shows
    decomposed instances, with red contours marking regions assigned to
    high-attention DIs. These examples are illustrative and do not constitute
    quantitative localization evaluation.
    }
    \label{fig:app_qualitative_attention}
\end{figure}

We further inspect highly attended instances from conventional tile-level MIL and DI-MIL. Figure~\ref{fig:app_qualitative_attention} shows examples across the four datasets. In these selected cases, decomposed instances tend to concentrate on more localized cellular content while excluding portions of surrounding background, staining variation, or artifacts. These examples are intended to illustrate the behavior of the learned attention and do not constitute quantitative localization evaluation.

\vspace{-8pt}
\vspace{-8pt}

\subsection{Paired Fold-level Statistics for Main Result}
\label{app:paired_stats}
To assess whether the mean improvements in Table~\ref{tab:main_results} are stable across matched folds, we additionally report fold-wise DI-minus-Tile differences. These analyses use the fold as the pairing unit and are intended to characterize effect consistency across the fixed cross-validation splits. Because training sets overlap across cross-validation folds, the resulting \(p\)-values should be interpreted as descriptive fold-paired evidence rather than as tests based on fully independent experimental replicates.
\vspace{-8pt}
\begin{table}[H]
\centering
\caption{
Fold-paired tests of Decomposed instances (DI) versus Tile on AUC.
Each comparison uses the same folds, features, splits, and MIL protocol as Table~\ref{tab:main_results}.
$\Delta$ denotes the mean fold-wise difference DI $-$ Tile in percentage points, and values in parentheses are $\mathrm{SD}_{\Delta}$, the standard deviation of the paired differences.
The 95\% CI is the paired Student's $t$ interval for the mean $\Delta$.
$p_{\mathrm{W}}$ is the two-sided exact Wilcoxon signed-rank $p$-value (zero differences dropped), and $p_t$ is the two-sided paired $t$-test $p$-value.
$n_{+}$ counts folds with $\Delta>0$.
The three institutional cohorts use 10 folds, whereas \textit{Oral2026} uses 5 folds.
For \textit{Oral2026}, the smallest attainable two-sided exact Wilcoxon $p$-value is 0.0625 when all five paired differences are non-zero.
}
\label{tab:main_paired_auc}
\scriptsize
\setlength{\tabcolsep}{4.0pt}
\renewcommand{\arraystretch}{1.08}

\begin{tabular}{llcccc}
\toprule
\multicolumn{6}{c}{\textbf{ABMIL}} \\
\cmidrule(lr){1-6}
Dataset
& $n$
& $\Delta$ ($\mathrm{SD}_{\Delta}$)
& 95\% CI
& $p_{\mathrm{W}}$ / $p_t$
& $n_{+}$ \\
\midrule

\multicolumn{6}{c}{\textbf{DINOv3}} \\
\midrule
Cervical
& 10
& \pdelta{0.12}{\scriptsize\,(1.66)}
& $[-1.07,\,1.31]$
& $0.695$ / $0.826$
& $6/10$ \\

\cmidrule(lr){1-6}
Serous Fluid
& 10
& \pdelta{0.42}{\scriptsize\,(2.12)}
& $[-1.09,\,1.93]$
& $0.695$ / $0.545$
& $6/10$ \\

\cmidrule(lr){1-6}
Respiratory
& 10
& \pdelta{1.38}{\scriptsize\,(4.61)}
& $[-1.92,\,4.68]$
& $0.695$ / $0.368$
& $5/10$ \\

\cmidrule(lr){1-6}
Oral2026
& 5
& \ndelta{2.34}{\scriptsize\,(1.92)}
& $[-4.72,\,0.04]$
& $0.062$ / $0.053$
& $0/5$ \\

\midrule
\multicolumn{6}{c}{\textbf{Smart-CCS}} \\
\midrule
Cervical
& 10
& \pdelta{0.33}{\scriptsize\,(2.22)}
& $[-1.26,\,1.91]$
& $0.557$ / $0.654$
& $6/10$ \\

\cmidrule(lr){1-6}
Serous Fluid
& 10
& \pdelta{3.41}{\scriptsize\,(1.59)}
& $[2.27,\,4.55]$
& $0.002^{**}$ / $<0.001^{***}$
& $10/10$ \\

\cmidrule(lr){1-6}
Respiratory
& 10
& \pdelta{8.14}{\scriptsize\,(3.56)}
& $[5.59,\,10.68]$
& $0.002^{**}$ / $<0.001^{***}$
& $10/10$ \\

\cmidrule(lr){1-6}
Oral2026
& 5
& \pdelta{0.26}{\scriptsize\,(2.28)}
& $[-2.57,\,3.08]$
& $0.812$ / $0.813$
& $3/5$ \\

\midrule
\multicolumn{6}{c}{\textbf{UniCAS}} \\
\midrule
Cervical
& 10
& \pdelta{0.67}{\scriptsize\,(1.90)}
& $[-0.69,\,2.03]$
& $0.375$ / $0.292$
& $6/10$ \\

\cmidrule(lr){1-6}
Serous Fluid
& 10
& \pdelta{1.09}{\scriptsize\,(2.95)}
& $[-1.02,\,3.20]$
& $0.232$ / $0.274$
& $7/10$ \\

\cmidrule(lr){1-6}
Respiratory
& 10
& \pdelta{4.85}{\scriptsize\,(5.05)}
& $[1.24,\,8.46]$
& $0.020^{*}$ / $0.014^{*}$
& $8/10$ \\

\cmidrule(lr){1-6}
Oral2026
& 5
& \pdelta{2.73}{\scriptsize\,(3.94)}
& $[-2.16,\,7.62]$
& $0.250$ / $0.196$
& $3/5$ \\

\bottomrule
\end{tabular}

\vspace{1.0em}

\begin{tabular}{llcccc}
\toprule
\multicolumn{6}{c}{\textbf{ACMIL}} \\
\cmidrule(lr){1-6}
Dataset
& $n$
& $\Delta$ ($\mathrm{SD}_{\Delta}$)
& 95\% CI
& $p_{\mathrm{W}}$ / $p_t$
& $n_{+}$ \\
\midrule

\multicolumn{6}{c}{\textbf{DINOv3}} \\
\midrule
Cervical
& 10
& \pdelta{0.03}{\scriptsize\,(2.74)}
& $[-1.93,\,1.99]$
& $1.00$ / $0.973$
& $5/10$ \\

\cmidrule(lr){1-6}
Serous Fluid
& 10
& \pdelta{0.68}{\scriptsize\,(2.90)}
& $[-1.39,\,2.76]$
& $0.625$ / $0.476$
& $6/10$ \\

\cmidrule(lr){1-6}
Respiratory
& 10
& \pdelta{0.96}{\scriptsize\,(1.97)}
& $[-0.46,\,2.37]$
& $0.232$ / $0.160$
& $5/10$ \\

\cmidrule(lr){1-6}
Oral2026
& 5
& \pdelta{0.99}{\scriptsize\,(2.17)}
& $[-1.70,\,3.68]$
& $0.438$ / $0.366$
& $3/5$ \\

\midrule
\multicolumn{6}{c}{\textbf{Smart-CCS}} \\
\midrule
Cervical
& 10
& \pdelta{1.59}{\scriptsize\,(2.72)}
& $[-0.36,\,3.53]$
& $0.160$ / $0.098$
& $6/10$ \\

\cmidrule(lr){1-6}
Serous Fluid
& 10
& \pdelta{3.30}{\scriptsize\,(2.29)}
& $[1.66,\,4.94]$
& $0.002^{**}$ / $0.001^{**}$
& $10/10$ \\

\cmidrule(lr){1-6}
Respiratory
& 10
& \pdelta{7.64}{\scriptsize\,(3.70)}
& $[5.00,\,10.29]$
& $0.004^{**}$ / $<0.001^{***}$
& $9/10$ \\

\cmidrule(lr){1-6}
Oral2026
& 5
& \pdelta{0.93}{\scriptsize\,(3.44)}
& $[-3.34,\,5.21]$
& $0.812$ / $0.577$
& $3/5$ \\

\midrule
\multicolumn{6}{c}{\textbf{UniCAS}} \\
\midrule
Cervical
& 10
& \pdelta{0.37}{\scriptsize\,(1.56)}
& $[-0.74,\,1.48]$
& $0.492$ / $0.472$
& $7/10$ \\

\cmidrule(lr){1-6}
Serous Fluid
& 10
& \pdelta{1.61}{\scriptsize\,(1.95)}
& $[0.21,\,3.01]$
& $0.037^{*}$ / $0.028^{*}$
& $8/10$ \\

\cmidrule(lr){1-6}
Respiratory
& 10
& \pdelta{2.02}{\scriptsize\,(2.56)}
& $[0.19,\,3.86]$
& $0.064$ / $0.034^{*}$
& $8/10$ \\

\cmidrule(lr){1-6}
Oral2026
& 5
& \pdelta{1.13}{\scriptsize\,(3.95)}
& $[-3.77,\,6.03]$
& $0.625$ / $0.556$
& $4/5$ \\

\bottomrule
\end{tabular}

\end{table}
\vspace{-4pt}
\vspace{-4pt}

\begin{table}[H]
    \centering
    \caption{
    Per-fold AUC $\Delta$ (DI $-$ Tile, percentage points) for every matched comparison in Table~\ref{tab:main_results}.
    The fold is the pairing unit.
    Med.\ $[Q_1,Q_3]$ summarises the fold-wise $\Delta$ distribution.
    }
    \label{tab:main_auc_fold_delta}
    \scriptsize
    \setlength{\tabcolsep}{2.4pt}
    \renewcommand{\arraystretch}{1.05}
    \begin{tabular}{ll|cccccccccc|c}
    \toprule
    Dataset & MIL
    & 0 & 1 & 2 & 3 & 4 & 5 & 6 & 7 & 8 & 9
    & Med.\ $[Q_1,Q_3]$ \\
    \midrule
    \multicolumn{13}{c}{\textbf{DINOv3}} \\
    \midrule
    \multirow{2}{*}{Cervical} & ABMIL & \pdelta{0.55} & \pdelta{0.42} & \pdelta{0.44} & \ndelta{1.05} & \pdelta{2.26} & \ndelta{0.36} & \pdelta{2.34} & \ndelta{2.81} & \ndelta{1.79} & \pdelta{1.20} & $+0.43\,[-0.87,\,+1.03]$ \\
     & ACMIL & \ndelta{3.85} & \ndelta{1.90} & \pdelta{4.31} & \pdelta{1.59} & \ndelta{3.22} & 0.00 & \pdelta{3.14} & \pdelta{2.05} & \ndelta{1.88} & \pdelta{0.06} & $+0.03\,[-1.90,\,+1.94]$ \\
    \cmidrule(lr){1-13}
    \multirow{2}{*}{Serous Fluid} & ABMIL & \pdelta{0.49} & \pdelta{3.64} & \ndelta{2.13} & \pdelta{1.03} & \pdelta{0.15} & \pdelta{4.19} & \ndelta{1.73} & \ndelta{1.49} & \pdelta{0.33} & \ndelta{0.27} & $+0.24\,[-1.19,\,+0.89]$ \\
     & ACMIL & \pdelta{3.92} & \pdelta{4.11} & \ndelta{2.16} & \pdelta{3.51} & \ndelta{1.27} & \pdelta{2.07} & \ndelta{4.23} & \ndelta{1.87} & \pdelta{1.11} & \pdelta{1.65} & $+1.38\,[-1.72,\,+3.15]$ \\
    \cmidrule(lr){1-13}
    \multirow{2}{*}{Respiratory} & ABMIL & \ndelta{2.58} & \ndelta{2.69} & \pdelta{2.53} & \ndelta{0.29} & \pdelta{9.15} & \pdelta{0.25} & \pdelta{5.92} & \pdelta{7.16} & \ndelta{4.31} & \ndelta{1.30} & $-0.02\,[-2.26,\,+5.07]$ \\
     & ACMIL & \pdelta{2.01} & \pdelta{3.56} & \ndelta{0.68} & \ndelta{0.02} & \pdelta{3.19} & \ndelta{0.95} & \ndelta{1.70} & \ndelta{0.77} & \pdelta{2.90} & \pdelta{2.03} & $+1.00\,[-0.75,\,+2.68]$ \\
    \cmidrule(lr){1-13}
    \multirow{2}{*}{Oral2026} & ABMIL & \ndelta{0.14} & \ndelta{1.13} & \ndelta{3.13} & \ndelta{5.13} & \ndelta{2.18} & -- & -- & -- & -- & -- & $-2.18\,[-3.13,\,-1.13]$ \\
     & ACMIL & \pdelta{0.41} & \pdelta{0.63} & \pdelta{4.77} & \ndelta{0.38} & \ndelta{0.48} & -- & -- & -- & -- & -- & $+0.41\,[-0.38,\,+0.63]$ \\
    \midrule
    \multicolumn{13}{c}{\textbf{Smart-CCS}} \\
    \midrule
    \multirow{2}{*}{Cervical} & ABMIL & \ndelta{0.88} & \ndelta{1.80} & \pdelta{4.22} & \ndelta{0.84} & \pdelta{1.42} & \pdelta{1.34} & \pdelta{1.42} & \pdelta{1.84} & \ndelta{3.70} & \pdelta{0.21} & $+0.78\,[-0.87,\,+1.42]$ \\
     & ACMIL & \ndelta{0.50} & \ndelta{1.80} & \pdelta{0.67} & \pdelta{3.26} & \pdelta{3.43} & \pdelta{3.51} & \pdelta{5.58} & \pdelta{4.28} & \ndelta{0.79} & \ndelta{1.77} & $+1.96\,[-0.72,\,+3.49]$ \\
    \cmidrule(lr){1-13}
    \multirow{2}{*}{Serous Fluid} & ABMIL & \pdelta{0.90} & \pdelta{2.79} & \pdelta{3.99} & \pdelta{4.58} & \pdelta{3.09} & \pdelta{2.65} & \pdelta{2.47} & \pdelta{4.54} & \pdelta{6.66} & \pdelta{2.44} & $+2.94\,[+2.51,\,+4.40]$ \\
     & ACMIL & \pdelta{0.70} & \pdelta{2.36} & \pdelta{7.18} & \pdelta{1.13} & \pdelta{2.67} & \pdelta{3.93} & \pdelta{0.29} & \pdelta{4.90} & \pdelta{5.95} & \pdelta{3.87} & $+3.27\,[+1.43,\,+4.66]$ \\
    \cmidrule(lr){1-13}
    \multirow{2}{*}{Respiratory} & ABMIL & \pdelta{7.19} & \pdelta{6.36} & \pdelta{13.19} & \pdelta{12.09} & \pdelta{7.56} & \pdelta{12.32} & \pdelta{3.64} & \pdelta{9.86} & \pdelta{5.76} & \pdelta{3.44} & $+7.37\,[+5.91,\,+11.53]$ \\
     & ACMIL & \pdelta{11.76} & \ndelta{1.06} & \pdelta{5.28} & \pdelta{9.67} & \pdelta{7.91} & \pdelta{11.43} & \pdelta{6.83} & \pdelta{9.94} & \pdelta{7.83} & \pdelta{6.87} & $+7.87\,[+6.84,\,+9.87]$ \\
    \cmidrule(lr){1-13}
    \multirow{2}{*}{Oral2026} & ABMIL & \pdelta{1.65} & \ndelta{2.63} & \ndelta{1.64} & \pdelta{2.69} & \pdelta{1.21} & -- & -- & -- & -- & -- & $+1.21\,[-1.64,\,+1.65]$ \\
     & ACMIL & \ndelta{1.38} & \pdelta{5.26} & \pdelta{1.34} & \ndelta{3.46} & \pdelta{2.91} & -- & -- & -- & -- & -- & $+1.34\,[-1.38,\,+2.91]$ \\
    \midrule
    \multicolumn{13}{c}{\textbf{UniCAS}} \\
    \midrule
    \multirow{2}{*}{Cervical} & ABMIL & \pdelta{0.33} & \pdelta{2.51} & \pdelta{0.75} & \pdelta{0.38} & \pdelta{4.22} & \ndelta{0.06} & \ndelta{1.38} & \pdelta{2.49} & \ndelta{1.88} & \ndelta{0.64} & $+0.36\,[-0.50,\,+2.06]$ \\
     & ACMIL & \pdelta{0.59} & \pdelta{1.80} & \pdelta{0.56} & \pdelta{1.25} & \ndelta{2.59} & \ndelta{1.51} & \pdelta{2.17} & \ndelta{0.67} & \pdelta{0.21} & \pdelta{1.88} & $+0.57\,[-0.45,\,+1.66]$ \\
    \cmidrule(lr){1-13}
    \multirow{2}{*}{Serous Fluid} & ABMIL & \pdelta{3.63} & \pdelta{0.32} & \ndelta{0.71} & \pdelta{2.94} & \pdelta{2.45} & \pdelta{2.27} & \ndelta{5.09} & \ndelta{1.33} & \pdelta{5.26} & \pdelta{1.13} & $+1.70\,[-0.45,\,+2.82]$ \\
     & ACMIL & \pdelta{2.01} & \pdelta{1.47} & \ndelta{1.25} & \pdelta{3.61} & \pdelta{0.92} & \pdelta{2.23} & \ndelta{1.99} & \pdelta{3.35} & \pdelta{1.89} & \pdelta{3.87} & $+1.95\,[+1.06,\,+3.07]$ \\
    \cmidrule(lr){1-13}
    \multirow{2}{*}{Respiratory} & ABMIL & \pdelta{11.35} & \pdelta{1.97} & \pdelta{3.58} & \pdelta{6.79} & \pdelta{5.22} & \ndelta{0.41} & \ndelta{1.78} & \pdelta{8.61} & \pdelta{0.23} & \pdelta{12.96} & $+4.40\,[+0.66,\,+8.16]$ \\
     & ACMIL & \pdelta{1.57} & \pdelta{2.94} & \pdelta{5.05} & \pdelta{1.01} & \pdelta{2.96} & \ndelta{2.11} & \ndelta{2.11} & \pdelta{4.97} & \pdelta{4.06} & \pdelta{1.90} & $+2.42\,[+1.15,\,+3.78]$ \\
    \cmidrule(lr){1-13}
    \multirow{2}{*}{Oral2026} & ABMIL & 0.00 & \pdelta{9.02} & \ndelta{0.45} & \pdelta{4.10} & \pdelta{0.97} & -- & -- & -- & -- & -- & $+0.97\,[+0.00,\,+4.10]$ \\
     & ACMIL & \pdelta{3.17} & \pdelta{2.51} & \pdelta{0.89} & \ndelta{5.51} & \pdelta{4.61} & -- & -- & -- & -- & -- & $+2.51\,[+0.89,\,+3.17]$ \\
    \bottomrule
    \end{tabular}
\end{table}
\vspace{-4pt}

\begin{table}[H]
    \centering
    \caption{
    Fold-paired tests of DI versus Tile on ACC and macro-F1, using the same matched folds as Table~\ref{tab:main_results}.
    ACC and macro-F1 are treated as secondary metrics and are reported for completeness; their fold-paired analyses should not be interpreted as independent replications of the AUC analysis.
    Mean $\Delta$, 95\% paired-$t$ CI, and two-sided exact Wilcoxon $p_{\mathrm{W}}$ follow Table~\ref{tab:main_paired_auc}.
    $^{*}\,p<0.05$, $^{**}\,p<0.01$, $^{***}\,p<0.001$.
    }
    \label{tab:main_paired_acc_f1}
    \scriptsize
    \setlength{\tabcolsep}{3.4pt}
    \renewcommand{\arraystretch}{1.08}
    \begin{tabular}{ll|ccc|ccc}
    \toprule
    \multirow{2}{*}{Dataset}
    & \multirow{2}{*}{Metric}
    & \multicolumn{3}{c|}{ABMIL}
    & \multicolumn{3}{c}{ACMIL} \\
    \cmidrule(lr){3-5}
    \cmidrule(lr){6-8}
    &
    & Mean $\Delta$ & 95\% CI & $p_{\mathrm{W}}$
    & Mean $\Delta$ & 95\% CI & $p_{\mathrm{W}}$ \\
    \midrule
    \multicolumn{8}{c}{\textbf{DINOv3}} \\
    \midrule
    \multirow{2}{*}{Cervical} & ACC & \pdelta{1.13} & $[-1.64,\,3.89]$ & $0.301$ & \pdelta{1.02} & $[-0.22,\,2.26]$ & $0.109$ \\
     & F1 & \pdelta{1.08} & $[-1.72,\,3.89]$ & $0.322$ & \pdelta{1.05} & $[-0.21,\,2.31]$ & $0.064$ \\
    \cmidrule(lr){1-8}
    \multirow{2}{*}{Serous Fluid} & ACC & \pdelta{0.77} & $[-0.96,\,2.50]$ & $0.275$ & \pdelta{0.96} & $[-1.48,\,3.39]$ & $0.426$ \\
     & F1 & \pdelta{0.79} & $[-0.97,\,2.56]$ & $0.232$ & \pdelta{0.95} & $[-1.49,\,3.39]$ & $0.557$ \\
    \cmidrule(lr){1-8}
    \multirow{2}{*}{Respiratory} & ACC & \pdelta{1.64} & $[-1.65,\,4.93]$ & $0.301$ & \pdelta{0.77} & $[-0.64,\,2.17]$ & $0.297$ \\
     & F1 & \pdelta{0.94} & $[-4.47,\,6.36]$ & $0.695$ & \pdelta{0.56} & $[-1.37,\,2.49]$ & $0.695$ \\
    \cmidrule(lr){1-8}
    \multirow{2}{*}{Oral2026} & ACC & \pdelta{2.77} & $[-0.50,\,6.04]$ & $0.125$ & \ndelta{0.86} & $[-4.46,\,2.73]$ & $0.875$ \\
     & F1 & \ndelta{4.67} & $[-14.46,\,5.11]$ & $0.312$ & \pdelta{1.02} & $[-7.27,\,9.31]$ & $1.00$ \\
    \midrule
    \multicolumn{8}{c}{\textbf{Smart-CCS}} \\
    \midrule
    \multirow{2}{*}{Cervical} & ACC & \pdelta{2.66} & $[0.80,\,4.52]$ & $0.023^{*}$ & \pdelta{3.16} & $[0.12,\,6.21]$ & $0.037^{*}$ \\
     & F1 & \pdelta{2.72} & $[0.78,\,4.66]$ & $0.027^{*}$ & \pdelta{3.26} & $[0.15,\,6.37]$ & $0.037^{*}$ \\
    \cmidrule(lr){1-8}
    \multirow{2}{*}{Serous Fluid} & ACC & \pdelta{3.54} & $[1.70,\,5.37]$ & $0.004^{**}$ & \pdelta{3.13} & $[1.37,\,4.90]$ & $0.004^{**}$ \\
     & F1 & \pdelta{3.49} & $[1.66,\,5.32]$ & $0.002^{**}$ & \pdelta{3.17} & $[1.40,\,4.93]$ & $0.004^{**}$ \\
    \cmidrule(lr){1-8}
    \multirow{2}{*}{Respiratory} & ACC & \pdelta{5.96} & $[4.13,\,7.79]$ & $0.002^{**}$ & \pdelta{6.15} & $[3.16,\,9.14]$ & $0.006^{**}$ \\
     & F1 & \pdelta{7.21} & $[4.23,\,10.18]$ & $0.002^{**}$ & \pdelta{8.25} & $[4.01,\,12.49]$ & $0.010^{**}$ \\
    \cmidrule(lr){1-8}
    \multirow{2}{*}{Oral2026} & ACC & \ndelta{1.30} & $[-5.49,\,2.90]$ & $0.625$ & \pdelta{2.80} & $[-1.13,\,6.72]$ & $0.188$ \\
     & F1 & \pdelta{4.07} & $[-11.68,\,19.83]$ & $0.812$ & \pdelta{3.43} & $[-3.15,\,10.02]$ & $0.188$ \\
    \midrule
    \multicolumn{8}{c}{\textbf{UniCAS}} \\
    \midrule
    \multirow{2}{*}{Cervical} & ACC & \pdelta{0.62} & $[-1.08,\,2.31]$ & $0.469$ & \pdelta{0.51} & $[-0.70,\,1.72]$ & $0.570$ \\
     & F1 & \pdelta{0.61} & $[-1.08,\,2.31]$ & $0.426$ & \pdelta{0.55} & $[-0.69,\,1.79]$ & $0.359$ \\
    \cmidrule(lr){1-8}
    \multirow{2}{*}{Serous Fluid} & ACC & \pdelta{2.25} & $[-0.25,\,4.74]$ & $0.098$ & \pdelta{1.91} & $[0.47,\,3.36]$ & $0.016^{*}$ \\
     & F1 & \pdelta{2.26} & $[-0.23,\,4.74]$ & $0.105$ & \pdelta{1.91} & $[0.46,\,3.37]$ & $0.020^{*}$ \\
    \cmidrule(lr){1-8}
    \multirow{2}{*}{Respiratory} & ACC & \pdelta{0.76} & $[-2.88,\,4.40]$ & $0.734$ & \pdelta{1.15} & $[-2.10,\,4.41]$ & $0.232$ \\
     & F1 & \pdelta{0.42} & $[-4.33,\,5.18]$ & $0.820$ & \pdelta{1.78} & $[-2.28,\,5.83]$ & $0.322$ \\
    \cmidrule(lr){1-8}
    \multirow{2}{*}{Oral2026} & ACC & \pdelta{0.89} & $[-9.33,\,11.10]$ & $0.625$ & \ndelta{0.22} & $[-3.97,\,3.52]$ & $1.00$ \\
     & F1 & \pdelta{5.41} & $[-2.24,\,13.05]$ & $0.125$ & \pdelta{2.75} & $[-4.85,\,10.35]$ & $0.438$ \\
    \bottomrule
    \end{tabular}
\end{table}
\vspace{-4pt}

\subsection{Detailed Cross-Validation Results for Representative MIL Baselines}
\label{app:detailed_results}
The compact main-text comparison reports method means to keep the three-backbone table readable.The following tables provide mean and standard deviation across cross-validation folds for methods for which fold-wise statistics are available in the supplied experimental results.

\begin{table}[ht]
\centering
\caption{
Detailed comparison with existing MIL methods on the \textit{Cervical} and \textit{Serous Fluid} datasets.
Results are reported as mean with standard deviation in parentheses.
Best mean results are shown in \textbf{bold}, and second-best mean results are \underline{underlined}.
}
\label{tab:appendix_cervical_serous}

\scriptsize
\setlength{\tabcolsep}{4pt}
\renewcommand{\arraystretch}{1.0}

\begin{tabular}{l|ccc|ccc}
\toprule
\multirow{2}{*}{Method}
& \multicolumn{3}{c|}{Cervical}
& \multicolumn{3}{c}{Serous Fluid} \\
\cmidrule(lr){2-4}
\cmidrule(lr){5-7}
& AUC & ACC & F1
& AUC & ACC & F1 \\
\midrule

\multicolumn{7}{c}{\textbf{DINOv3}} \\
\midrule

ABMIL
& \res{91.40}{2.11}
& \res{84.56}{3.45}
& \res{84.40}{3.44}
& \res{85.37}{2.57}
& \res{78.76}{2.23}
& \res{78.70}{2.27} \\

ACMIL
& \res{\underline{93.83}}{1.96}
& \res{\underline{87.22}}{3.33}
& \res{\underline{87.01}}{3.43}
& \res{\underline{86.11}}{3.49}
& \res{79.13}{3.21}
& \res{79.11}{3.21} \\

DTFD
& \res{92.29}{2.72}
& \res{86.00}{3.49}
& \res{85.66}{3.68}
& \res{85.13}{3.13}
& \res{78.06}{2.42}
& \res{77.99}{2.41} \\

ASMIL
& \res{92.57}{1.79}
& \res{87.02}{3.33}
& \res{86.89}{3.40}
& \res{84.31}{3.01}
& \res{77.40}{3.43}
& \res{77.32}{3.52} \\

DSMIL
& \res{90.00}{2.49}
& \res{83.84}{1.53}
& \res{83.59}{1.55}
& \res{83.06}{2.86}
& \res{76.52}{3.06}
& \res{76.46}{3.10} \\

TransMIL
& \res{91.32}{1.98}
& \res{83.44}{3.24}
& \res{83.21}{3.32}
& \res{85.03}{3.03}
& \res{77.40}{3.20}
& \res{77.35}{3.22} \\

MambaMIL
& \res{90.94}{2.73}
& \res{86.71}{3.61}
& \res{86.48}{3.76}
& \res{86.00}{2.37}
& \res{\underline{79.24}}{1.95}
& \res{\underline{79.19}}{1.97} \\

HMIL
& \res{91.93}{2.52}
& \res{84.15}{4.02}
& \res{83.84}{4.30}
& \res{84.78}{2.82}
& \res{77.14}{2.64}
& \res{77.03}{2.61} \\

PTCMIL
& \res{91.86}{2.11}
& \res{84.66}{2.99}
& \res{84.44}{3.02}
& \res{85.82}{2.92}
& \res{78.50}{2.80}
& \res{78.44}{2.85} \\

\greyrow
DI-ACMIL (Ours)
& \res{\textbf{93.86}}{1.97}
& \res{\textbf{88.24}}{3.12}
& \res{\textbf{88.06}}{3.23}
& \res{\textbf{86.79}}{3.50}
& \res{\textbf{80.09}}{3.56}
& \res{\textbf{80.06}}{3.56} \\

\midrule
\multicolumn{7}{c}{\textbf{Smart-CCS}} \\
\midrule

ABMIL
& \res{94.38}{1.58}
& \res{87.42}{2.41}
& \res{87.28}{2.53}
& \res{84.22}{2.22}
& \res{76.52}{2.74}
& \res{76.49}{2.74} \\

ACMIL
& \res{94.31}{1.90}
& \res{86.82}{3.36}
& \res{86.61}{3.48}
& \res{\underline{85.04}}{2.82}
& \res{\underline{77.51}}{3.20}
& \res{\underline{77.43}}{3.25} \\

DTFD
& \res{92.37}{1.69}
& \res{86.71}{3.03}
& \res{86.55}{2.99}
& \res{83.91}{3.03}
& \res{76.30}{2.77}
& \res{76.28}{2.77} \\

ASMIL
& \res{91.20}{2.38}
& \res{86.61}{2.25}
& \res{86.48}{2.34}
& \res{78.77}{3.99}
& \res{74.35}{2.37}
& \res{74.32}{2.39} \\

DSMIL
& \res{93.90}{2.30}
& \res{87.22}{3.32}
& \res{87.10}{3.35}
& \res{81.88}{2.71}
& \res{73.76}{2.83}
& \res{73.68}{2.90} \\

TransMIL
& \res{93.30}{1.32}
& \res{87.12}{3.40}
& \res{86.95}{3.57}
& \res{83.17}{3.25}
& \res{75.45}{2.73}
& \res{75.36}{2.75} \\

MambaMIL
& \res{92.92}{2.76}
& \res{\underline{87.74}}{3.52}
& \res{\underline{87.65}}{3.58}
& \res{83.00}{2.13}
& \res{75.52}{2.43}
& \res{75.46}{2.46} \\

HMIL
& \res{93.65}{0.80}
& \res{87.12}{2.16}
& \res{87.04}{2.22}
& \res{83.39}{2.59}
& \res{75.52}{3.14}
& \res{75.44}{3.17} \\

PTCMIL
& \res{\underline{94.61}}{1.18}
& \res{87.02}{4.12}
& \res{86.82}{4.38}
& \res{84.87}{3.20}
& \res{76.59}{3.73}
& \res{76.46}{3.86} \\

\greyrow
DI-ACMIL (Ours)
& \res{\textbf{95.89}}{1.21}
& \res{\textbf{89.98}}{3.36}
& \res{\textbf{89.87}}{3.35}
& \res{\textbf{88.33}}{2.09}
& \res{\textbf{80.64}}{2.62}
& \res{\textbf{80.60}}{2.63} \\

\midrule
\multicolumn{7}{c}{\textbf{UniCAS}} \\
\midrule

ABMIL
& \res{93.68}{2.43}
& \res{88.66}{2.36}
& \res{88.54}{2.39}
& \res{84.04}{2.21}
& \res{\underline{76.85}}{2.04}
& \res{\underline{76.81}}{2.01} \\

ACMIL
& \res{\underline{95.42}}{0.74}
& \res{\underline{89.06}}{1.73}
& \res{\underline{88.84}}{1.85}
& \res{\underline{84.45}}{2.40}
& \res{76.52}{2.27}
& \res{76.48}{2.26} \\

DTFD
& \res{92.66}{2.59}
& \res{86.30}{3.12}
& \res{86.09}{3.19}
& \res{82.76}{1.51}
& \res{74.97}{1.44}
& \res{74.93}{1.44} \\

ASMIL
& \res{90.58}{2.98}
& \res{87.43}{1.72}
& \res{87.30}{1.80}
& \res{80.63}{2.01}
& \res{74.31}{2.00}
& \res{74.28}{2.01} \\

DSMIL
& \res{92.17}{2.26}
& \res{86.20}{3.39}
& \res{85.98}{3.49}
& \res{80.31}{3.48}
& \res{73.17}{2.54}
& \res{73.08}{2.48} \\

TransMIL
& \res{92.06}{2.41}
& \res{84.97}{2.44}
& \res{84.78}{2.43}
& \res{80.18}{2.60}
& \res{73.17}{2.58}
& \res{73.12}{2.61} \\

MambaMIL
& \res{93.34}{1.93}
& \res{86.51}{2.79}
& \res{86.31}{2.82}
& \res{83.12}{2.32}
& \res{76.22}{2.00}
& \res{76.15}{2.03} \\

HMIL
& \res{93.55}{2.65}
& \res{87.02}{3.06}
& \res{86.91}{3.06}
& \res{82.04}{3.27}
& \res{75.08}{3.05}
& \res{74.95}{3.13} \\

PTCMIL
& \res{94.68}{1.64}
& \res{87.73}{3.49}
& \res{87.54}{3.70}
& \res{83.68}{2.53}
& \res{74.05}{2.94}
& \res{73.83}{3.04} \\

\greyrow
DI-ACMIL (Ours)
& \res{\textbf{95.79}}{1.62}
& \res{\textbf{89.57}}{2.14}
& \res{\textbf{89.39}}{2.23}
& \res{\textbf{86.06}}{2.41}
& \res{\textbf{78.43}}{2.04}
& \res{\textbf{78.39}}{2.01} \\

\bottomrule
\end{tabular}
\end{table}

\begin{table}[ht]
\centering
\caption{
Detailed comparison with existing MIL methods on the \textit{Respiratory} and \textit{Oral2026} datasets.
Results are reported as mean with standard deviation in parentheses.
Best mean results are shown in \textbf{bold}, and second-best mean results are \underline{underlined}.
}
\label{tab:appendix_respiratory_oral}

\scriptsize
\setlength{\tabcolsep}{4pt}
\renewcommand{\arraystretch}{1.05}

\begin{tabular}{l|ccc|ccc}
\toprule
\multirow{2}{*}{Method}
& \multicolumn{3}{c|}{Respiratory}
& \multicolumn{3}{c}{Oral2026} \\
\cmidrule(lr){2-4}
\cmidrule(lr){5-7}
& AUC & ACC & F1
& AUC & ACC & F1 \\
\midrule

\multicolumn{7}{c}{\textbf{DINOv3}} \\
\midrule

ABMIL
& \res{79.38}{4.18}
& \res{74.73}{4.14}
& \res{70.64}{5.31}
& \res{88.06}{5.07}
& \res{79.84}{2.85}
& \res{70.93}{5.04} \\

ACMIL
& \res{\underline{81.05}}{4.66}
& \res{\underline{75.98}}{3.15}
& \res{\underline{72.11}}{3.87}
& \res{\underline{88.37}}{5.15}
& \res{80.77}{2.89}
& \res{71.45}{5.90} \\

DTFD
& \res{75.77}{6.15}
& \res{71.57}{5.40}
& \res{66.90}{5.63}
& \res{78.98}{14.35}
& \res{81.50}{2.93}
& \res{58.29}{14.02} \\

ASMIL
& \res{73.11}{8.67}
& \res{74.54}{4.51}
& \res{69.95}{6.47}
& \res{85.02}{6.08}
& \res{80.65}{3.23}
& \res{58.73}{12.90} \\

DSMIL
& \res{74.42}{5.23}
& \res{72.43}{3.21}
& \res{68.16}{3.98}
& \res{85.60}{5.74}
& \res{\underline{82.67}}{3.92}
& \res{\textbf{72.57}}{4.25} \\

TransMIL
& \res{75.79}{6.36}
& \res{71.47}{4.57}
& \res{66.50}{6.57}
& \res{79.52}{8.52}
& \res{72.53}{7.49}
& \res{63.87}{6.74} \\

MambaMIL
& \res{80.38}{2.54}
& \res{74.06}{2.46}
& \res{70.34}{2.57}
& \res{83.22}{8.74}
& \res{\textbf{82.90}}{2.91}
& \res{63.85}{11.98} \\

HMIL
& \res{79.19}{4.08}
& \res{74.16}{4.13}
& \res{70.28}{4.51}
& \res{82.16}{4.77}
& \res{81.51}{5.82}
& \res{65.58}{12.18} \\

PTCMIL
& \res{79.64}{5.12}
& \res{74.16}{3.73}
& \res{68.46}{5.46}
& \res{86.14}{5.42}
& \res{80.60}{4.93}
& \res{64.33}{11.68} \\

\greyrow
DI-ACMIL (Ours)
& \res{\textbf{82.01}}{4.92}
& \res{\textbf{76.75}}{4.19}
& \res{\textbf{72.67}}{5.70}
& \res{\textbf{89.36}}{4.53}
& \res{79.91}{4.09}
& \res{\underline{72.47}}{5.20} \\

\midrule
\multicolumn{7}{c}{\textbf{Smart-CCS}} \\
\midrule

ABMIL
& \res{79.36}{4.13}
& \res{74.16}{3.11}
& \res{70.04}{4.38}
& \res{83.91}{5.38}
& \res{83.18}{2.74}
& \res{67.18}{13.01} \\

ACMIL
& \res{\underline{81.21}}{3.45}
& \res{\underline{76.17}}{3.19}
& \res{\underline{71.83}}{4.47}
& \res{\underline{88.20}}{5.44}
& \res{81.73}{6.14}
& \res{72.26}{4.25} \\

DTFD
& \res{76.38}{5.15}
& \res{71.37}{4.93}
& \res{67.48}{5.31}
& \res{82.19}{10.55}
& \res{83.16}{2.72}
& \res{51.88}{14.73} \\

ASMIL
& \res{72.91}{5.59}
& \res{73.39}{3.62}
& \res{68.93}{5.32}
& \res{78.26}{6.09}
& \res{81.79}{4.41}
& \res{65.19}{11.63} \\

DSMIL
& \res{75.55}{5.27}
& \res{70.60}{5.38}
& \res{65.80}{6.25}
& \res{84.65}{5.31}
& \res{80.92}{8.60}
& \res{70.49}{6.48} \\

TransMIL
& \res{75.85}{5.09}
& \res{70.99}{4.85}
& \res{66.04}{6.32}
& \res{86.54}{7.64}
& \res{76.92}{2.92}
& \res{68.15}{5.16} \\

MambaMIL
& \res{79.16}{5.30}
& \res{74.07}{4.37}
& \res{70.11}{4.92}
& \res{84.32}{6.30}
& \res{83.61}{5.85}
& \res{67.57}{9.90} \\

HMIL
& \res{78.06}{3.71}
& \res{73.39}{4.08}
& \res{69.91}{4.42}
& \res{84.71}{2.73}
& \res{84.05}{4.59}
& \res{73.48}{6.74} \\

PTCMIL
& \res{79.13}{4.41}
& \res{73.39}{3.21}
& \res{69.00}{4.73}
& \res{87.79}{6.26}
& \res{\underline{84.08}}{4.48}
& \res{\underline{73.75}}{6.52} \\

\greyrow
DI-ACMIL (Ours)
& \res{\textbf{88.85}}{2.07}
& \res{\textbf{82.32}}{3.10}
& \res{\textbf{80.08}}{3.48}
& \res{\textbf{89.13}}{4.86}
& \res{\textbf{84.52}}{5.25}
& \res{\textbf{75.69}}{8.27} \\

\midrule
\multicolumn{7}{c}{\textbf{UniCAS}} \\
\midrule

ABMIL
& \res{75.92}{4.99}
& \res{73.78}{3.39}
& \res{\underline{69.89}}{4.28}
& \res{85.92}{4.96}
& \res{\underline{85.30}}{7.65}
& \res{71.74}{8.80} \\

ACMIL
& \res{\underline{78.70}}{2.84}
& \res{\underline{74.54}}{2.70}
& \res{69.63}{3.91}
& \res{\underline{87.62}}{3.40}
& \res{84.82}{1.27}
& \res{71.63}{6.63} \\

DTFD
& \res{70.90}{6.49}
& \res{68.30}{3.76}
& \res{63.07}{4.60}
& \res{85.06}{5.16}
& \res{\textbf{86.10}}{4.42}
& \res{\underline{72.42}}{9.78} \\

ASMIL
& \res{72.07}{4.77}
& \res{72.14}{4.14}
& \res{67.04}{5.71}
& \res{85.73}{3.21}
& \res{82.05}{2.85}
& \res{65.13}{10.62} \\

DSMIL
& \res{71.99}{6.03}
& \res{71.09}{5.57}
& \res{64.88}{8.34}
& \res{82.39}{2.39}
& \res{83.39}{5.42}
& \res{69.75}{5.38} \\

TransMIL
& \res{70.33}{6.07}
& \res{68.11}{6.71}
& \res{60.96}{8.32}
& \res{86.52}{3.37}
& \res{84.25}{2.65}
& \res{64.57}{12.20} \\

MambaMIL
& \res{76.02}{4.20}
& \res{71.18}{3.58}
& \res{65.98}{4.97}
& \res{83.22}{5.97}
& \res{82.20}{7.97}
& \res{70.03}{10.74} \\

HMIL
& \res{74.07}{5.96}
& \res{71.75}{4.60}
& \res{68.71}{4.98}
& \res{85.21}{2.45}
& \res{82.80}{4.95}
& \res{68.86}{1.68} \\

PTCMIL
& \res{76.25}{4.46}
& \res{72.14}{3.91}
& \res{65.70}{5.77}
& \res{86.31}{3.42}
& \res{83.71}{3.02}
& \res{70.24}{5.93} \\

\greyrow
DI-ACMIL (Ours)
& \res{\textbf{80.73}}{4.05}
& \res{\textbf{75.70}}{4.04}
& \res{\textbf{71.41}}{4.46}
& \res{\textbf{88.76}}{3.34}
& \res{84.60}{2.40}
& \res{\textbf{74.38}}{5.66} \\

\bottomrule
\end{tabular}
\end{table}

\clearpage

\subsection{Cost Analysis by Storage}
\label{app:storage_cost}
To complement the relative resource comparison in Figure \ref{fig:tile_cost}, we report the absolute feature-storage requirements measured on the Oral2026 dataset, comprising 363 WSIs. All methods use 1024-dimensional Smart-CCS embeddings stored in float16 format. We distinguish between the storage required by the feature instances actually consumed by the MIL aggregator and the physical size of the cached feature files.
\paragraph{Storage of MIL input features.}
Table~\ref{tab:storage_cost} reports the total number of MIL instances and their corresponding feature-storage requirements across different tile sizes. Conventional tile-based MIL retains one global [CLS] embedding per source tile, whereas DI-MIL constructs $k=4$ decomposed instances from each original 2048-pixel tile. Each stored instance occupies $1024 \times 2 = 2048$ bytes.

\begin{table}[H]
\centering
\caption{Absolute MIL feature-storage requirements on Oral2026 using Smart-CCS and ACMIL. Feature storage accounts for the float16 embeddings actually consumed by the MIL aggregator, excluding file-format overhead and auxiliary metadata. DI-MIL uses the original 2048-pixel tiles with $k=4$. Performance is reported as mean AUC and macro-F1 (\%).}
\label{tab:storage_cost}
\small
\begin{tabular}{lrrrr}
\toprule
Method & Instances & Storage (GiB) & AUC & Macro-F1 \\
\midrule
Tile 2048 & 0.94M & 1.79 & 88.20 & 72.26 \\
Tile 1200 & 2.67M & 5.09 & 88.95 & 73.89 \\
DI-MIL ($k=4$) & 3.75M & 7.14 & 89.13 & 75.69 \\
Tile 600 & 10.40M & 19.83 & 88.73 & 74.40 \\
Tile 300 & 40.52M & 77.30 & 86.20 & 74.35 \\
\bottomrule
\end{tabular}
\end{table}

DI-MIL requires 7.14 GiB of MIL input features, compared with 1.79 GiB for the original Tile 2048 baseline. This fourfold increase reflects the expansion of the MIL bag from $N$ to $4N$, rather than additional image extraction or encoder forward passes. Compared with Tile 1200, DI-MIL requires approximately 2.05 GiB of additional feature storage while improving macro-F1 by 1.80 percentage points. In contrast, direct fine-grid tiling at 600 and 300 pixels requires 19.83 and 77.30 GiB, respectively, while yielding lower AUC and macro-F1 than DI-MIL.

\paragraph{Physical storage of cached features.}
The actual cached feature files contain additional information beyond the embeddings consumed by MIL. Specifically, our feature-extraction pipeline stores one [CLS] token and four cluster embeddings per tile, together with auxiliary cluster-assignment maps. The same 2048-pixel feature cache is shared by conventional tile-based MIL and DI-MIL; the latter selects the four cluster embeddings instead of the [CLS] embedding and therefore does not require a separate copy of the source features.

Across all 363 WSIs, the 2048-pixel feature cache occupies 9.17 GiB on disk, including 8.93 GiB of stored token embeddings and 0.22 GiB of cluster-assignment maps. The corresponding feature caches for 1200-, 600-, and 300-pixel tiling occupy 26.13, 101.80, and 396.74 GiB, respectively. These physical file sizes include four cluster embeddings per tile that are not used by the conventional single-instance baselines. Therefore, the MIL input-feature sizes in Table~\ref{tab:storage_cost} provide a more representative comparison of the feature storage required by each instance-construction strategy, whereas the physical file sizes reflect our shared experimental caching format.

\paragraph{Why not directly use all dense tokens as MIL instances?}
A natural alternative to DI-MIL is to treat every dense spatial token produced by the frozen encoder as an independent MIL instance. However, this would substantially increase the storage and computational requirements of slide-level aggregation. As reported in Table~7, Smart-CCS produces 256 dense spatial tokens per encoded tile. Directly exposing all tokens to the MIL aggregator would therefore expand the slide-level bag from $N$ to $256N$ instances, compared with only $4N$ instances in DI-MIL.

For Oral2026, retaining all dense tokens as float16, 1024-dimensional MIL instances would require approximately 456.96 GiB of feature storage, compared with 7.14 GiB for DI-MIL. This represents a 64-fold increase in both the number of MIL instances and their feature-storage requirements, despite using exactly the same source tiles and encoder forward passes. The actual downstream memory footprint would be further affected by intermediate activations, attention computations, and model-specific training requirements.

Importantly, DI-MIL performs token clustering independently within each source tile and caches only the resulting four instance embeddings. In contrast, directly using all dense tokens would propagate the complete token sequence into slide-level learning, substantially increasing feature storage, data-loading overhead, and the computational burden of downstream aggregation. This cost becomes particularly consequential for large-scale cytopathology cohorts containing thousands of WSIs.

These results motivate decomposed-instance construction as a practical intermediate strategy: DI-MIL preserves multiple independently weighted sub-tile representations while reducing the number of instances exposed to the MIL aggregator by a factor of 64 relative to direct dense-token aggregation.

\subsection{Cost Analysis by Computational Time}
\label{app:cost_analysis_by_time}

While the main text evaluates cost efficiency from a storage perspective, here we analyze the empirical runtime requirements (GPU-hours) of DI-MIL compared against conventional fine-grid tiling strategies on the Respiratory dataset ($n = 1041$ slides).

\begin{figure}[H]
    \centering
    \includegraphics[width=1.0\linewidth]{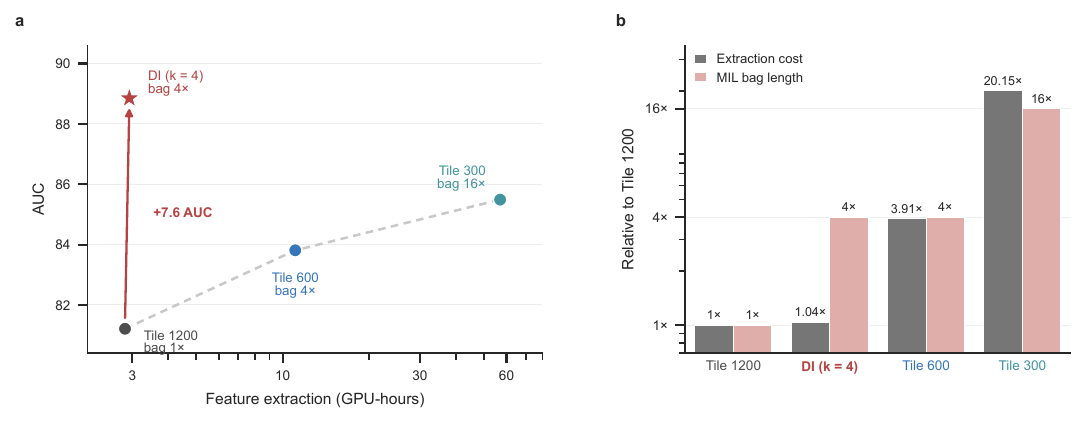}
    \caption{\textbf{Feature-extraction cost, MIL bag length, and AUC on Respiratory.} \textbf{(a)} AUC versus measured CCS feature-extraction GPU-hours for conventional tiles (dashed) and DI ($ k =4 $). Marker annotations give the median MIL bag relative to Tile 1200. \textbf{(b)} The same methods, split into extraction cost and MIL bag length, both relative to Tile 1200. DI matches the 4× bag of Tile 600 while remaining at Tile-1200 extraction cost. $n = 1041$ slides; MIL times are isolated ACMIL steps on a RTX 3090 GPU and are not taken from the contended training wall-clock.}
    \label{fig:rcs_di_vs_fine_tile_cost}
\end{figure}

As shown in Figure~\ref{fig:rcs_di_vs_fine_tile_cost}b, physically reducing tile size introduces severe computational bottlenecks during feature extraction. Compared to the baseline Tile 1200 ($1\times$), cropping finer tiles escalates extraction runtime drastically: Tile 600 requires $3.91\times$ GPU-hours, while Tile 300 inflates feature extraction time by $20.15\times$ ($\sim 58.4$ GPU-hours). In contrast, DI-MIL ($k=4$) performs lightweight token clustering in feature space post-encoding, incurring a negligible total runtime penalty of only $1.04\times$ the baseline extraction cost.

Furthermore, DI-MIL achieves a superior trade-off between diagnostic performance and computational efficiency (Figure~\ref{fig:rcs_di_vs_fine_tile_cost}a). While fine-grid tiling yields modest AUC gains at exponential computational expense ($85.5\%$ AUC at $20.15\times$ compute for Tile 300), DI-MIL ($k=4$) achieves a significantly higher AUC of $88.8\%$ ($+7.6\%$ over Tile 1200). It matches the $4\times$ bag length of Tile 600 while maintaining virtually the same extraction cost as Tile 1200. This confirms that DI-MIL provides Pareto-dominant efficiency, eliminating the heavy GPU forward-pass overhead required by traditional sub-tiling.

\end{document}